\documentclass[12pt,cleveref]{jmlr}
\usepackage{times}
\usepackage[T1]{fontenc}
\usepackage{microtype,booktabs,longtable,array}
\setcitestyle{citesep={,}}
\jmlrproceedings{}{}
\numberwithin{equation}{section}
\title[Visible Tails]{Visible Tails: Characterizing Adaptation to Unknown Moments in Heavy-Tailed Bandits}
\author[Do and Nguyen]{%
\Name[N.T. Do]{Do Ngoc Tuan} \Email{tuan.dongoc@phenikaa-uni.edu.vn}\\
\addr Phenikaa School of Information Technology, Phenikaa University, Hanoi 12116, Vietnam
\AND
\Name[D.H. Nguyen]{Nguyen Duc Hai} \Email{duchai001@e.ntu.edu.sg}\\
\addr College of Computing and Data Science, Nanyang Technological University, Singapore%
}
\begin{document}
\maketitle
\begin{abstract}
Heavy-tailed bandit learners must tell a poor arm from an arm whose favorable rewards have not yet appeared. We characterize this obstruction through a scale of one-sided visibility conditions on the optimal reward law: the mean hidden in a rare event is at most a constant multiple of a retained central moment. A quantile-trimmed UCB index that is given the visibility constants attains arm-specific moment rates at every exponent, with no lower bound on the exponent. Conversely, every strategy with a fixed exploration budget has polynomial regret on each law that violates the corresponding condition. For the variance benchmark we determine the order of both budget constants in the visibility level. The frontier depends on the largest exponent the benchmark promises, and the resulting classes are strictly nested. The same index has logarithmic expected regret when rewards are integrable, with a coefficient that can be exponential in the inverse gap.

\end{abstract}
\section{Introduction}\label{sec:intro}

A stochastic bandit learner must stay optimistic about rewards it has not yet seen. Heavy tails make this hard, because the rare large rewards that carry much of an arm's mean can stay hidden for a long time. When rewards have a moment of order \(1+\varepsilon\in(1,2]\) bounded by \(u\), robust UCB indices that know \((\varepsilon,u)\) achieve regret of order \(u^{1/\varepsilon}\Delta^{-1/\varepsilon}\log T\) on an arm with mean gap \(\Delta\) over \(T\) rounds \citep{bubeck2013}. Without these parameters, a run of small observations has two explanations: a poor arm, or the bulk of an excellent arm whose large rewards have not appeared. \citet{genalti2024} turn this ambiguity into lower bounds showing that the known-parameter rate is out of reach with no further assumption, and they recover that rate by data-driven trimming under truncated non-positivity of the optimal arm. Symmetry supports a second adaptive guarantee \citep{tamas2025}. These conditions are mutually incomparable and both exclude Pareto rewards, which led \citet{genalti2025} to ask, as part of a COLT open problem, whether some assumption is better than the others or encompasses all distributions for which adaptation comes at no additional cost.

This paper addresses that question. For a reward set \(E\) of probability \(1-p\), let \(b_E\) be the centered mean omitted by \(E\) and let \(v_E=(1-p)\operatorname{Var}[X\mid E]\). A law has \emph{variance-core visibility}, written \(\mathrm{VCV}(a,q)\) (\Cref{def:vcv}), if
\[
b_E\le a\sqrt{p\,v_E}
\]
for every set \(E\) with omitted probability \(0<p\le q\). The condition is a bias-variance balance. After about \(1/p\) pulls the omitted event has typically not occurred, so the sample looks like draws from the law conditioned on \(E\), whose mean has standard error about \(\sqrt{p\operatorname{Var}[X\mid E]}\). Visibility says that the mean still unseen is at most a constant multiple of this standard error, so a confidence width built from the observed spread also covers the unseen upside. The condition is one-sided, because hidden upside alone harms optimism, and it is imposed on an optimal arm alone. It is also translation invariant, whereas a common shift of all rewards inflates raw moments without bound. We measure a strategy by its \emph{exploration budget} \((C_0,C_1)\): against a deterministic optimal arm it pulls a competitor with standard deviation \(\sigma\) and gap \(\Delta\) at most \(\log T\,[C_0+(C_1\sigma/\Delta)^2]\) times. The allowance \(C_0\) pays for ruling out a competitor of zero variance, and has the form of the forced-exploration term of \citet{genalti2024}. The scale constant \(C_1\) multiplies the known-parameter rate. Central moments of order \(1+\varepsilon\) define budgets at every exponent in the same way.

\subsection{Our results}

\textbf{The frontier.} \Cref{thm:frontier} has three parts. VT-UCB (visible-tail UCB), a quantile-trimmed UCB index with an independent pilot center, receives \((a,q)\) and the horizon. It has variance budget \((40+50/q,\,4000(1+a))\) and attains the same bound against every competitor, at every exponent \(\varepsilon\in(0,1]\), whenever the optimal law is in \(\mathrm{VCV}(a,q)\). Conversely, a strategy with variance budget \((C_0,C_1)\) has regret of order \(T^{1-\eta}\), with an explicit \(\eta<1\), on each law outside \(\mathrm{VCV}(4C_1,\min\{1/2,1/(8C_0)\})\). This converse concerns one strategy and one law, and it gives polynomial regret where a class-level statement gives a larger constant. Finally, serving \(\mathrm{VCV}(a,q)\) forces \(C_0\ge1/\log\frac1{1-q}\) and \(C_1\ge\sqrt{a^2+2}\), so the optimal allowance is of order \(1/q\) and the optimal scale constant is of order \(1+a\). \Cref{cor:class} restates the frontier for classes of optimal laws.

Two observations drive the proof. Because the retained second moment of \(E\) around any point is at least \(v_E\), visibility holds at every center at once, so an inaccurate pilot on the optimal arm leaves its index optimistic and the pilot needs accuracy on the competitor alone. In the other direction, conditioning a law on \(E\) produces a competitor at KL divergence \(-\log(1-p)\), which takes about \(1/p\) pulls to distinguish from the original law and whose variance is exactly what the budget charges. The same quantity \(v_E\) therefore governs the upper and the lower bound.

\textbf{The visibility scale.} Replacing the retained variance by a retained central moment of order \(1+\varepsilon\) gives classes \(\mathcal V_\varepsilon(a,q)\), with \(\mathrm{VCV}=\mathcal V_1\). \Cref{thm:scale} shows that these classes grow strictly as \(\varepsilon\) decreases. A benchmark that promises rates up to a largest exponent \(\bar\varepsilon\) has frontier \(\mathcal V_{\bar\varepsilon}\), with no dependence on a smallest exponent (\Cref{cor:scale}). Inside the single class \(\mathcal V_\varepsilon(1,1/4)\) with \(\varepsilon<1\), every variance-budget strategy has regret \(T^{1-\delta}\) on some law for each \(\delta>0\), while VT-UCB at exponent \(\varepsilon\) has logarithmic regret throughout. In the raw-moment setting of the open problem, \Cref{app:raw} gives the achievability, converse and allowance bound for the variance budget, with threshold visibility at the origin. That raw class contains every truncated non-positive law, every law symmetric about its mean, and the translated Pareto laws with tail index bounded away from one. One condition therefore covers the assumptions known to support adaptation together with the Pareto rewards highlighted by \citet{genalti2025}.

\textbf{Logarithmic regret from a first moment.} \Cref{thm:integrability} proves instance-dependent \(O(\log T)\) regret for VT-UCB on every instance with a visible optimal arm and integrable competitors. Every arm, including the optimal one, needs a finite first moment and nothing more. A competitor with a central moment of order \(1+\varepsilon\) contributes the rate of its own exponent and scale, so the index adapts arm by arm to heterogeneous unknown exponents. The analysis separates the two errors of a trimmed mean, the bias from discarding large observations and the sampling fluctuation of the rest. Both vanish for every integrable law as the trimmed fraction tends to zero, so every positive gap is resolved at a finite sample size. \Cref{cor:endpoint} gives the endpoint rate \(\Delta^{-1/(\alpha-1)}\log T\) for polynomial tail envelopes, including Pareto competitors with infinite \(\alpha\)th moment.

\textbf{Instance-level sharpness.} Adding a rare favorable atom to a visible competitor law keeps the law visible (\Cref{lem:contamination}), which turns the retained variances of a competitor into a lower bound on its pulls (\Cref{thm:instance}). For a finite-variance competitor the bound is of order \(a^2\sigma^2/\Delta^2\), and for a polynomial tail of index \(\alpha\) it is of order \((a/\Delta)^{\alpha/(\alpha-1)}\), so the visibility constant enters the lower bound with the same power as in the upper bound. The coefficient in \Cref{thm:integrability} admits no polynomial bound in the inverse gap. \Cref{thm:gap} determines its order on an explicit family with tail \(1/[t(\log t)^2]\), which has no finite moment of order above one: the coefficient of VT-UCB and the information lower bound for consistent strategies are both of order \(\Delta^{-1}e^{A/\Delta}\), and their ratio is the explicit universal constant stated after the theorem.

\textbf{Unknown visibility.} No strategy has logarithmic regret on every instance with a visible optimal arm, and inputs that grow with time lose a factor diverging arbitrarily slowly (\Cref{prop:consequences}).

\subsection{Related work}\label{sec:related}

\textbf{Adaptation to unknown moments.} The lower bounds of \citet{genalti2024} frame the problem, and truncated non-positivity of the optimal arm supports their AdaR-UCB guarantee. Optimal-arm tail-sign assumptions also support adaptation in \citet{huang2022} and \citet{chen2025}. \citet{genalti2025} observe that AdaR-UCB keeps tight bounds on Pareto rewards, which violate the assumption, because the trimming bias has the order of the variance term. Visibility turns this observation into a condition. Raw visibility contains truncated non-positivity, and raw VT-UCB with the universal inputs \((1/2,1/2)\) attains the raw scale benchmark on that class. Variance-core visibility and truncated non-positivity are incomparable (\Cref{app:examples}), which reflects the role of location. \citet{tamas2025} construct the parameter-free RMM-UCB under symmetry of every arm about its mean, with gap-dependent \(\log^2 T\) regret. Every integrable symmetric law lies in \(\mathrm{VCV}(\sqrt{3/2},1/4)\), so VT-UCB with these universal inputs achieves \(O(\log T)\) regret using symmetry of the optimal arm alone. \citet{genalti2026} resolve the assumption-free part of this open problem and state that its remaining part is to identify the weakest additional assumption that recovers the known-parameter rates. Our characterization addresses that remaining part. Their trade-off theorem and our converse both rest on an under-exploration change of measure, and our contribution is the law-specific visibility geometry and the quantitative budget.

\textbf{Mean estimation and confidence widths.} Robust mean estimation \citep{catoni2012, devroye2016, lugosi2019} supplies point estimators with sharp deviations, and a bandit index additionally requires a computable width. \citet{oliveira2025} prove finite-sample guarantees for the order-statistic trimmed mean under central-moment conditions, with a trimming level set by the confidence. \citet{hammouda2026} characterize unknown-variance confidence intervals through an effective sample size and prove that adapting to the scale defining their classes is impossible, with sample-size-dependent classes for two-sided inference. Our condition is fixed in the sample size, controls the downward bias that matters for optimism, and characterizes bandit regret. These analyses share the classical difficulty of unseen mean mass \citep{bahadur1956}. VT-UCB combines trimming and pilot centering with the empirical-variance optimism of \citet{audibert2009} and \citet{maurer2009}, and its distinctive feature is a width that remains valid at every pilot center.

\textbf{Regret under a first moment.} \citet{ashutosh2021} obtain arbitrarily slow superlogarithmic expected regret over laws having some higher moment, using distribution-oblivious algorithms. \citet{gupta2026} give almost-sure guarantees for integrable rewards, and their expected-regret theorem uses sub-exponential tails. \Cref{thm:integrability} gives expected \(O(\log T)\) regret for integrable arms given the visibility constants, and \Cref{thm:gap} determines the coefficient of this guarantee up to a constant factor on an explicit family. With unknown visibility, the rate of \Cref{prop:consequences} is of the same kind as these forced-exploration rates. Our converse uses change of measure and reverse-KL alternatives \citep{lai1985, kaufmann2016, agrawal2021}. For rewards bounded above by a known endpoint, \citet{honda2010,honda2015} express the minimum divergence to a higher-mean law through the dual \(\max_\lambda\mathbb E[\log(1-\lambda(X-\mu))]\), which is the information quantity in our converse. Their IMED regret theorem assumes a moment generating function near the origin on an optimal arm and uses the endpoint as an input. It therefore applies to the instances of \Cref{sec:gap}, whose optimal arm is a point mass, and attains there the coefficient of our lower bound. VT-UCB uses neither the endpoint nor a moment generating function, and its guarantee extends to every visible optimal arm.

\textbf{Other structure.} Other work adapts to an unknown reward range \citep{hadiji2023} or uses a known kurtosis bound for scale-free regret \citep{lattimore2017}. Pareto-specific Thompson sampling exploits a parametric family \citep{lee2023}, whereas our learner receives no tail formula and no tail parameter. Adversarial heavy-tailed bandits \citep{cheng2024} and the comparison of heavy tails with contamination \citep{cherapanamjeri2025} address adversarially generated rewards and corrupted data.
\section{Model, visibility and budget}\label{sec:model}

There are \(K\ge2\) arms and a known horizon \(T\ge3\). Pulling arm \(i\) returns an independent reward of law \(\nu_i\) on \(\mathbb R\), with \(\mathbb E_{\nu_i}[|X|]<\infty\), and an instance is the tuple \(\boldsymbol\nu=(\nu_1,\ldots,\nu_K)\). Write \(\mu_i=\mathbb E_{\nu_i}[X]\), assume a unique optimal arm \(i^\star\), put \(\Delta_i=\mu_{i^\star}-\mu_i\), and let \(N_i(T)\) count the pulls of arm \(i\) in \(T\) rounds. Expected regret is
\[
R_T=\sum_{i\ne i^\star}\Delta_i\mathbb E[N_i(T)].
\]
Unless explicitly stated, no moment above order one is assumed. We write \(\delta_x\) for the point mass at \(x\).

For \(0<\varepsilon\le1\) put \(\gamma_\varepsilon=\varepsilon/(1+\varepsilon)\), and let
\[
\sigma_\varepsilon(\nu)=\bigl(\mathbb E_\nu[|X-\mu|^{1+\varepsilon}]\bigr)^{1/(1+\varepsilon)}\in[0,\infty]
\]
be the central scale of a law \(\nu\) with mean \(\mu\), with \(\sigma_{i,\varepsilon}=\sigma_\varepsilon(\nu_i)\). Estimating a mean to accuracy \(\Delta\) at confidence \(1/T\) under a finite scale \(\sigma_\varepsilon\) takes about \(\log T\,(\sigma_\varepsilon/\Delta)^{1/\gamma_\varepsilon}\) samples, which is the pull count of robust UCB with known parameters \citep{bubeck2013}.

Fix a law of \(X\) with mean \(\mu\). For a measurable reward set \(E\subseteq\mathbb R\) of positive probability, put
\[
\begin{aligned}
p&=\mathbb P[X\notin E],&\mu_E&=\mathbb E[X\mid X\in E],\\
b_E&=\mathbb E[(X-\mu)\mathbf1_{E^c}],&
\sigma_{E,\varepsilon}&=\bigl(\mathbb E[|X-\mu_E|^{1+\varepsilon}\mathbf1_E]\bigr)^{1/(1+\varepsilon)}.
\end{aligned}
\]
Thus \(b_E=(1-p)(\mu-\mu_E)\) measures how far the retained mean falls below the true mean, and \(\sigma_{E,\varepsilon}\) measures the spread of the retained rewards, with \(v_E=\sigma_{E,1}^2=(1-p)\operatorname{Var}[X\mid E]\). An atom is either retained or omitted whole.

\begin{definition}[Moment-core visibility]\label{def:vcv}
Let \(a\ge0\), \(0<q\le1/2\) and \(0<\varepsilon\le1\). An integrable law belongs to \(\mathcal V_\varepsilon(a,q)\) if
\[
b_E\le a\,p^{\gamma_\varepsilon}\sigma_{E,\varepsilon}
\qquad(0<p\le q,\ \sigma_{E,\varepsilon}<\infty).
\]
The case \(\varepsilon=1\) is variance-core visibility, \(\mathrm{VCV}(a,q)=\mathcal V_1(a,q)\), which reads \(b_E\le a\sqrt{pv_E}\).
\end{definition}

When \(\sigma_{E,\varepsilon}=0\), the condition reads \(b_E\le0\). Each class is invariant under translations and positive rescaling, grows with \(a\), shrinks with \(q\), and contains every point mass. The constant \(a\) bounds the unseen mean in standard errors at sample size \(1/p\), and \(q\) is the largest omitted probability covered. The learner's inputs are \((a,q)\), the exponent of the class and the horizon.

\begin{definition}[Scale budget]\label{def:budget}
A strategy has budget \((C_0,C_1)\) at exponent \(\varepsilon\) if, for every \(T\ge3\) and every two-armed instance \((\delta_\theta,\pi)\) in which \(\theta\) exceeds the mean \(\mu_\pi\) of \(\pi\) and \(\sigma_\varepsilon(\pi)<\infty\),
\[
\mathbb E[N_\pi(T)]\le\log T\left[C_0+\left(\frac{C_1\sigma_\varepsilon(\pi)}{\theta-\mu_\pi}\right)^{1/\gamma_\varepsilon}\right].
\]
The variance budget is the case \(\varepsilon=1\).
\end{definition}

For a largest promised exponent \(\bar\varepsilon\in(0,1]\), the scale benchmark is
\[
\mathcal B_T(C_0,C_1;\bar\varepsilon)=\log T\sum_{i\ne i^\star}\Delta_i
\left[C_0+\inf_{0<\varepsilon\le\bar\varepsilon}
\left(\frac{C_1\sigma_{i,\varepsilon}}{\Delta_i}\right)^{1/\gamma_\varepsilon}\right].
\]
A strategy with \(R_T\le\mathcal B_T(C_0,C_1;\bar\varepsilon)\) on every instance with a deterministic optimal arm has budget \((C_0,C_1)\) at exponent \(\bar\varepsilon\). The allowance \(C_0\) pays for ruling out a competitor of zero scale and for initialization, and \(C_1\) multiplies the known-parameter rate. The constant sits inside the power, as in robust UCB, and \Cref{cor:instance} shows that the visibility constant must enter as \(a^{1/\gamma_\varepsilon}\). The benchmark needs no lower bound on the exponent.

\section{The frontier}\label{sec:frontier}

A strategy is \emph{uniformly efficient} on an instance if \(R_T=o(T^\delta)\) for every \(\delta>0\). For a set \(E\) with omitted probability \(p\), write \(d=-\log(1-p)\), which is the KL divergence from the law conditioned on \(E\) to the law itself.

\begin{theorem}[Variance frontier]\label{thm:frontier}
Let \(a\ge0\) and \(0<q\le1/2\).

\textup{(i) Achievability.} If the optimal law is in \(\mathrm{VCV}(a,q)\) and every arm is integrable, VT-UCB with inputs \((a,q)\) satisfies \(R_T\le\mathcal B_T(40+50/q,\,4000(1+a);1)\) for every \(K\) and \(T\ge3\). In particular it has variance budget \((40+50/q,\,4000(1+a))\).

\textup{(ii) Law-by-law converse.} Let a strategy have variance budget \((C_0,C_1)\). Let \(\nu\) be an integrable law and \(E\) a set with \(0<p\le\min\{1/2,1/(8C_0)\}\), \(v_E<\infty\) and \(b_E>4C_1\sqrt{pv_E}\), where \(1/(8C_0)=+\infty\) if \(C_0=0\) here and below. Put \(\Delta=b_E/[2(1-p)]\), \(\theta=(\mu+\mu_E)/2\) and
\[
\eta=2d\bigl[C_0+4C_1^2(1-p)v_E/b_E^2\bigr].
\]
Then \(\eta<1\) and
\[
R_T(\delta_\theta,\nu)\ge\frac\Delta4\,T^{1-\eta}\qquad\text{whenever }2\eta\log T\le dT.
\]

\textup{(iii) Exchange.} If a strategy has variance budget \((C_0,C_1)\) and is uniformly efficient on every two-armed instance whose optimal law is in \(\mathrm{VCV}(a,q)\), then
\[
C_0\ge\frac1{\log\frac1{1-q}}\ge\frac{1-q}q\qquad\text{and}\qquad C_1\ge\sqrt{a^2+2}.
\]
\end{theorem}

\begin{proof}[Proof of (ii)]
Let \(\pi=\nu(\cdot\mid E)\) and compare \(\mathcal J_0=(\delta_\theta,\pi)\) with \(\mathcal J_1=(\delta_\theta,\nu)\), writing \(\mathbb P_0,\mathbb P_1\) for the laws of the interaction. Since \(\mu-\mu_E=b_E/(1-p)>0\), the point mass is optimal in \(\mathcal J_0\) and \(\nu\) is optimal in \(\mathcal J_1\), each with gap \(\Delta\). The law \(\pi\) has variance \(v_E/(1-p)\), so the budget gives \(\mathbb E_0[N_2(T)]\le n_T:=\eta\log T/(2d)\). Let \(F_T=\{N_2(T)\le2n_T\}\), so that \(\mathbb P_0[F_T]\ge1/2\) by Markov's inequality, and let \(\mathcal G_T\) be the event that every observed reward of arm 2 lies in \(E\). On \(E\) the law \(\nu\) has density \(1-p\) with respect to \(\pi\), and the strategy acts identically under both instances, so
\[
\mathbb P_1[F_T\cap\mathcal G_T]=\mathbb E_0\bigl[\mathbf1_{F_T}(1-p)^{N_2(T)}\bigr]\ge\tfrac12(1-p)^{2n_T}=\tfrac12T^{-\eta}.
\]
When \(2\eta\log T\le dT\), the event \(F_T\) implies \(N_2(T)\le T/2\). Arm 1 is suboptimal in \(\mathcal J_1\), hence \(R_T(\mathcal J_1)=\Delta\,\mathbb E_1[T-N_2(T)]\ge\frac{\Delta T}2\mathbb P_1[F_T]\ge\frac\Delta4T^{1-\eta}\). For the exponent, \(d\le p/(1-p)\) gives \(2C_0d\le4C_0p\le1/2\), and the violation \(v_E/b_E^2<1/(16C_1^2p)\) gives \(8C_1^2(1-p)v_Ed/b_E^2<(1-p)d/(2p)\le1/2\).
\end{proof}

Part (i) is the case \(\bar\varepsilon=1\) of \Cref{thm:integrability}. Part (iii) is proved in \Cref{app:exchange}. The allowance bound compares \(\delta_0\) with Bernoulli laws of success probability just above \(q\). The scale bound perturbs a two-point competitor of variance \(\sigma^2\) by a small tilt and a rare favorable atom that keep it in \(\mathrm{VCV}(a,q)\), which raises the mean by \(\Delta\) at KL divergence \(\Delta^2/[\sigma^2(a^2+2)]\). Parts (i) and (ii) describe the same geometry up to a change of constants in \((a,q)\). With (iii), the optimal allowance on \(\mathrm{VCV}(a,q)\) is of order \(1/q\) and the optimal scale constant is of order \(1+a\).

\begin{corollary}[Class frontier]\label{cor:class}
Fix \(\bar\varepsilon\in(0,1]\) and let \(\mathcal D\) be a class of integrable laws containing every point mass. Some strategy and finite \((C_0,C_1)\) satisfy \(R_T\le\mathcal B_T(C_0,C_1;\bar\varepsilon)\) for all \(T\ge3\) on every two-armed instance whose optimal law is in \(\mathcal D\) if and only if \(\mathcal D\subseteq\mathcal V_{\bar\varepsilon}(a,q)\) for some finite \(a\) and positive \(q\). The first property implies \(\mathcal D\subseteq\mathcal V_{\bar\varepsilon}(4C_1,\min\{1/2,1/(8C_0)\})\), and the second gives \((C_0,C_1)=(40+50/q,\,8000(1+a))\) for every number of arms.
\end{corollary}

Sufficiency is \Cref{thm:integrability}, and necessity applies the converse at exponent \(\bar\varepsilon\) (\Cref{thm:converse}) to each law of \(\mathcal D\). For the variance budget, the raw-moment formulation has the same achievability, converse and allowance bound, with threshold visibility at the origin in place of the variance core and raw scales in place of central ones (\Cref{app:raw}). The two frontiers differ: a two-atom family is uniformly raw visible and lies outside every class \(\mathrm{VCV}(a,q)\), because the raw scale pays for a large artificial location.

\section{One learner for every exponent}\label{sec:learner}

At confidence \(1/T\), a sample of size \(n\) is uninformative about events of probability below about \(\log T/n\). VT-UCB therefore discards a fixed number \(\ell\) of the largest deviations and builds its index from the rest, so the trimmed fraction \(r=\ell/n\) shrinks as the estimate reaches further into the tail.

Set the trimming count \(\ell=\lceil24\log T\rceil\), the pilot size \(m=2\lceil16\log T\rceil+1\), and the warm-up \(n_0=\lceil2\ell/q\rceil\), which keeps \(2r\le q\). Pull each arm \(m\) times and reserve these observations solely for its median \(Z_i\). Collect \(n_0\) fresh observations per arm. If the horizon ends during initialization, stop there. Let \(X_{ij}\) be the \(j\)th fresh observation of arm \(i\). At fresh sample size \(n\), let \(S_i\) index the \(n-\ell\) observations with smallest \(|X_{ij}-Z_i|\), resolving ties by observation index. The trimmed mean, retained second moment, largest retained deviation and width are
\[
\begin{aligned}
\widehat\mu_i&=\frac{\sum_{j\in S_i}X_{ij}}{n-\ell},&
\widehat V_i&=\frac{\sum_{j\in S_i}(X_{ij}-Z_i)^2}{n-\ell},\\
\widehat M_i&=\max_{j\in S_i}|X_{ij}-Z_i|,&
\widehat w_i&=\sqrt{2\widehat V_i\ell/n}+9\ell\widehat M_i/n.
\end{aligned}
\]
Centered VT-UCB selects an arm maximizing \(U_i=\widehat\mu_i+3(1+a)\widehat w_i\), using fixed arm order for ties. The term \(3\widehat w_i\) covers sampling error and the term \(3a\widehat w_i\) covers the mean removed by trimming, which visibility bounds by a multiple of the width. Shifting or positively rescaling all rewards leaves the actions unchanged, and the pilot and fresh samples are independent by construction. For a class \(\mathcal V_{\bar\varepsilon}(a,q)\) with \(\bar\varepsilon<1\), the index uses the retained moment of order \(1+\bar\varepsilon\) in place of the retained second moment:
\[
\widehat V_i^{(\bar\varepsilon)}=\frac{\sum_{j\in S_i}|X_{ij}-Z_i|^{1+\bar\varepsilon}}{n-\ell},\ \
\widehat w_i^{(\bar\varepsilon)}=\Bigl(\frac\ell n\Bigr)^{\gamma_{\bar\varepsilon}}\bigl(\widehat V_i^{(\bar\varepsilon)}\bigr)^{\frac1{1+\bar\varepsilon}}+\frac{9\ell\widehat M_i}n,\ \
U_i^{(\bar\varepsilon)}=\widehat\mu_i+8(1+a)\widehat w_i^{(\bar\varepsilon)}.
\]

\begin{theorem}[VT-UCB at every exponent]\label{thm:integrability}
Let \(\bar\varepsilon\in(0,1]\). Suppose the optimal law is in \(\mathcal V_{\bar\varepsilon}(a,q)\) and every arm is integrable, and run VT-UCB with the index \(U_i\) if \(\bar\varepsilon=1\) and \(U_i^{(\bar\varepsilon)}\) otherwise. For every suboptimal arm \(i\) and every \(\varepsilon\in(0,\bar\varepsilon]\) with \(\sigma_{i,\varepsilon}<\infty\),
\[
\mathbb E[N_i(T)]\le m+n_0+1+\frac{18}{T^2}+\frac1{T^3}
+\ell\left(\frac{800(1+a)\sigma_{i,\varepsilon}}{\Delta_i}\right)^{1/\gamma_\varepsilon}
\]
when \(\bar\varepsilon=1\), and the same bound holds with \(1600\) in place of \(800\) when \(\bar\varepsilon<1\). Consequently \(R_T\le\mathcal B_T(40+50/q,\,4000(1+a);1)\), respectively \(R_T\le\mathcal B_T(40+50/q,\,8000(1+a);\bar\varepsilon)\). Moreover \(\mathbb E[N_i(T)]=O(\log T)\) for every integrable competitor.
\end{theorem}

Each exponent bounds the same expectation, so the infimum in the benchmark needs no union bound and no exponent floor, and the optimal arm needs no moment above one. The proof is in \Cref{app:profiles}. For the index \(U_i\), all statements rest on one function of the trimming level \(r\), the profile \(H\), which is the smallest gap the index can resolve at that level. For a law \(\nu\) on \(\mathbb R\) with mean \(\bar\mu\) and \(Y\sim\nu\), write \(p(M)=\mathbb P[|Y|>M]\), \(V(M)=\mathbb E[Y^2\mathbf1_{\{|Y|\le M\}}]\) and \(s(M)=\sqrt{p(M)V(M)}+Mp(M)\), and define, for \(0<r\le1/4\),
\[
\begin{aligned}
M_-(r)&=\inf\{M\ge0:p(M)\le2r\},&M_+(r)&=\inf\{M\ge0:p(M)\le r/2\},\\
B_\nu(r)&=(\mathbb E[Y\mid |Y|\le M_-(r)]-\bar\mu)^+,&
W_\nu(r)&=3\sqrt{rV(M_+(r))}+11rM_+(r),\\
H_{\nu,a}(r)&=B_\nu(r)+3(2+a)W_\nu(r).&&
\end{aligned}
\]
Here \(M_\pm\) are population quantiles that bracket the empirical threshold, \(B_\nu\) is the amount by which trimming can inflate an estimated mean, and \(W_\nu\) bounds the width. Given all pilots and writing \(\nu_i-Z_i\) for the law of \(X-Z_i\), on a concentration event arm \(i\) is pulled at fresh sample size \(n\) only when \(\Delta_i\le H_{\nu_i-Z_i,a}(\ell/n)\). For each competitor, this event fails with probability at most \(14/T^3\) over all sample sizes.
The concentration event comes from \Cref{lem:trimming}. Binomial bounds place the empirical threshold \(\widehat M\) between \(M_-(r)\) and \(M_+(r)\). Bernstein bounds at \(M_-\) then show that \(\widehat w\) dominates the trimmed-mean error and \(s(M_-)\), which gives optimism, and a Bernstein bound at \(M_+\) gives \(\widehat w\le W_\nu(r)\), which controls competitors.

For integrable laws, \(rM_+(r)\to0\) and \(rV(M_+(r))\to0\). Absolute continuity of the integral makes the omitted mean vanish too. \Cref{app:finite-means} makes this convergence uniform over \(|Z_i-\mu_i|\le4\mathbb E[|X-\mu_i|]\), an event of probability at least \(1-T^{-4}\). Every positive gap is therefore resolved at a finite fresh sample size, which gives logarithmic regret. That sample size depends on the law through its profile.

\begin{corollary}[Polynomial endpoint]\label{cor:endpoint}
Under the assumptions of \Cref{thm:integrability} with \(\bar\varepsilon=1\), if a suboptimal law satisfies \(\mathbb P[|X-\mu_i|>t]\le A_i t^{-\alpha_i}\) for all \(t>0\), where \(1<\alpha_i<2\), its expected pull count \(\mathbb E[N_i(T)]\) is at most
\[
m+n_0+1+14/T^2+T^{-3}
+\ell\left(\frac{D_{\alpha_i}(a)(16A_i)^{1/\alpha_i}}{\Delta_i}\right)^{\alpha_i/(\alpha_i-1)},
\]
with explicit \(D_\alpha(a)\) in \Cref{app:finite-means}. The parameters \(A_i,\alpha_i\) are unknown to the learner. This includes Pareto endpoints with infinite \(\alpha_i\)th moment, and \Cref{cor:instance} gives a lower bound with the same power of \(a/\Delta_i\) for competitors with a polynomial-tail core.
\end{corollary}

\section{The visibility scale}\label{sec:scale}

\begin{theorem}[Nesting and separation]\label{thm:scale}
\textup{(i)} For every \(0<\varepsilon\le1\), \(a\ge0\) and \(0<q\le1/2\),
\[
\mathrm{VCV}(a,q)\subseteq\mathcal V_\varepsilon\bigl(\tfrac32(\sqrt2\,a+1),\,q/2\bigr).
\]
\textup{(ii)} Fix \(0<\varepsilon<1\) and, for \(0<p\le1/4\), let
\[
\nu_p=\tfrac{1-p}2\,\delta_{-1}+\tfrac{1-p}2\,\delta_1+p\,\delta_{x_p},\qquad x_p=p^{-1/(1+\varepsilon)}.
\]
Then \(\nu_p\in\mathcal V_\varepsilon(1,1/4)\) and \(\mathbb E_{\nu_p}[|X|^{1+\varepsilon}]\le2\) for every \(p\), and no class \(\mathrm{VCV}(a,q)\) contains the whole family.
\end{theorem}

For (i), a set \(E\) is cut at the level \(t=\sigma_{E,\varepsilon}p^{-1/(1+\varepsilon)}\), which omits at most a further probability \(p\) and leaves a retained variance of at most \(t^{1-\varepsilon}\sigma_{E,\varepsilon}^{1+\varepsilon}\). Variance-core visibility of the cut set then gives the claim (\Cref{app:scale}). For (ii), each atom at \(\pm1\) has mass at least \(3/8\), so the only set with omitted probability in \((0,1/4]\) omits the atom at \(x_p\). For it \(\mu_E=0\), \(\sigma_{E,\varepsilon}=(1-p)^{1/(1+\varepsilon)}\) and \(b_E=p(1-p)x_p=(1-p)p^{\gamma_\varepsilon}\). Hence \(b_E/(p^{\gamma_\varepsilon}\sigma_{E,\varepsilon})=(1-p)^{\gamma_\varepsilon}\le1\), while \(b_E/\sqrt{pv_E}=\sqrt{1-p}\,p^{\gamma_\varepsilon-1/2}\to\infty\) as \(p\to0\). The moment equals \((1-p)+p\,x_p^{1+\varepsilon}\le2\).

\begin{corollary}[The largest promised exponent sets the frontier]\label{cor:scale}
\textup{(i)} For a benchmark that promises every exponent in \((0,\bar\varepsilon]\), the frontier is \(\mathcal V_{\bar\varepsilon}\), by \Cref{cor:class}. Promising further smaller exponents changes nothing.

\textup{(ii)} Let \(0<\varepsilon<1\) and let a strategy have variance budget \((C_0,C_1)\). On the family of \Cref{thm:scale}, for all small \(p\), the bound of \Cref{thm:frontier}\textup{(ii)} holds with
\[
\eta=2\bigl(C_0+4C_1^2p^{-2\gamma_\varepsilon}\bigr)\log\tfrac1{1-p}\longrightarrow0\qquad(p\to0).
\]
Hence for every \(\delta>0\) some law of the single class \(\mathcal V_\varepsilon(1,1/4)\) forces regret of order \(T^{1-\delta}\), while VT-UCB with the index \(U_i^{(\varepsilon)}\) and inputs \((1,1/4)\) has logarithmic regret on the whole class.
\end{corollary}

Part (ii) uses the set that omits \(x_p\), for which \((1-p)v_E/b_E^2=p^{-2\gamma_\varepsilon}\) and \(2\gamma_\varepsilon<1\).

\section{Instance-level sharpness}\label{sec:gap}

We now bound from below the pulls of a given competitor. For a law \(\pi\), a level \(0<q\le1/4\) and a weight \(\varpi>0\), define the contamination capacity
\[
\Gamma_\pi(\varpi)=\inf\Bigl\{\sqrt{\bigl(\varpi+\pi(F^c)\bigr)v_F}\;:\;\pi(F^c)\le q\Bigr\},
\]
where \(v_F\) is computed under \(\pi\) and \(F=\mathbb R\) is allowed.

\begin{lemma}[Visible contamination]\label{lem:contamination}
Let \(0<q\le1/4\), \(0\le a'\le a\), \(\pi\in\mathrm{VCV}(a',2q)\) and \(0<\varpi\le q/2\). If \(0\le\iota\le(a-a')\Gamma_\pi(\varpi)\), then the law \(\nu=(1-\varpi)\pi+\varpi\,\delta_{\mu_\pi+\iota/\varpi}\) belongs to \(\mathrm{VCV}(a,q)\), has mean \(\mu_\pi+\iota\), and satisfies \(\operatorname{KL}(\pi\Vert\nu)\le\log\frac1{1-\varpi}\).
\end{lemma}

\begin{theorem}[Instance lower bound]\label{thm:instance}
Let \(0<q\le1/4\) and \(0\le a'\le a\). Consider an instance whose optimal law is in \(\mathrm{VCV}(a,q)\) and in which a competitor has law \(\pi\in\mathrm{VCV}(a',2q)\) and gap \(\Delta\), and a strategy that is uniformly efficient on every instance whose optimal law is in \(\mathrm{VCV}(a,q)\). For every \(\varpi\in(0,q/2]\) with \((a-a')\Gamma_\pi(\varpi)>\Delta\),
\[
\liminf_{T\to\infty}\frac{\mathbb E[N_\pi(T)]}{\log T}\ge\frac1{\log\frac1{1-\varpi}}.
\]
\end{theorem}

Indeed, for \(\Delta<\iota<(a-a')\Gamma_\pi(\varpi)\) the law \(\nu\) of \Cref{lem:contamination} is visible and has mean above the optimal mean. Replacing the competitor by \(\nu\) gives an instance on which the strategy is uniformly efficient, and the change of measure of \Cref{lem:information} applies.

\begin{corollary}[Finite variance and polynomial tails]\label{cor:instance}
In the setting of \Cref{thm:instance} let \(a>a'\).

\textup{(i)} If \(\underline v_q=\inf\{v_F:\pi(F^c)\le q\}>0\) and \(8\Delta^2\le(a-a')^2\underline v_q\,q\), the lower limit is at least \((a-a')^2\underline v_q/(8\Delta^2)\). If \(\pi\) has finite variance, it is at least \((a-a')^2\sigma_1(\pi)^2(1+o(1))/\Delta^2\) as \(\Delta\to0\).

\textup{(ii)} Let \(1<\alpha<2\) and suppose \(v_F\ge\underline v_\alpha\,\pi(F^c)^{-(2-\alpha)/\alpha}\) whenever \(0<\pi(F^c)\le q\), with \(v_{\mathbb R}=\infty\). Then for all small \(\Delta\) the lower limit is at least
\[
\frac12\left(\frac{(a-a')\sqrt{\underline v_\alpha}}{2\Delta}\right)^{\alpha/(\alpha-1)}.
\]
\end{corollary}

With \(\varepsilon=\alpha-1\) the exponent in (ii) is \(1/\gamma_\varepsilon\), so the visibility constant enters the lower bound as \(a^{1/\gamma_\varepsilon}\), as in \Cref{cor:endpoint}. \Cref{app:contamination} contains the proofs and a Pareto mixture that satisfies (ii) with \(a'=0\). The next family shows how large the coefficient can be without any higher moment.

Fix \(A=e/10\). Let \(\xi\) have mass \(9/10\) at zero and survival probability \(A/[y(\log y)^2]\) for \(y\ge e\), constant \(1/10\) on \([0,e)\). Then \(\mathbb E[\xi]=2A\) and \(\mathbb E[\xi^{1+\varepsilon}]=\infty\) for every \(\varepsilon>0\). For \(0<\Delta<A/2\), put \(x_\Delta=2A-\Delta\) and define
\[
P_\Delta=\mathcal L(x_\Delta-\xi),\qquad
\mathcal I_\Delta=(\delta_0,P_\Delta),
\]
where \(\mathcal L\) denotes the law of its argument. The second arm has mean \(-\Delta\), yet it usually pays \(x_\Delta>0\), more than the optimal arm, and its low mean comes from rare large losses. An omitted reward set of mass at most \(1/4\) cannot contain the positive atom, so it removes only rewards below the mean, and both arms lie in the fixed class \(\mathrm{VCV}(0,1/4)\). The loss mass beyond the quantile at tail probability \(r\) is about \(A/\log(1/r)\), so the sample must reach \(r\approx e^{-A/\Delta}\) before the gap surfaces.

\begin{theorem}[Exponential gap complexity]\label{thm:gap}
Universal constants \(C_\xi,\Delta_\xi>0\) exist such that centered VT-UCB with \((a,q)=(0,1/4)\) satisfies
\[
\limsup_{T\to\infty}\frac{R_T(\mathcal I_\Delta)}{\log T}
\le C_\xi\Delta^{-1}e^{A/\Delta}\qquad(0<\Delta<\Delta_\xi).
\]
Every strategy with logarithmic regret on all integrable two-arm instances in \(\mathrm{VCV}(0,1/4)\) satisfies
\[
\liminf_{T\to\infty}\frac{R_T(\mathcal I_\Delta)}{\log T}
\ge\frac{\Delta}{d_\Delta},\qquad
d_\Delta\sim\frac{\Delta^2}{Ae}e^{-A/\Delta}.
\]
Here \(d_\Delta=\max_{0\le\lambda<1/x_\Delta}\mathbb E_{P_\Delta}[\log(1-\lambda X)]\). It suffices to assume logarithmic regret on \(\mathcal I_\Delta\) and the explicit finite-variance alternatives in \Cref{app:gap}. Limits in \(T\) precede the small-gap limit.
\end{theorem}

The prefactors are \(48Ae^{\kappa+1}\) for the upper bound and \(Ae\) for the lower bound, with \(\kappa=133+18\sqrt2\). The two bounds therefore have the same order, and their ratio is the explicit proof constant \(48e^\kappa\approx3\times10^{70}\).

With high probability the pilot median is the positive atom, so the centered rewards have law \(-\xi\). Writing \(H=H_{-\xi,0}\) for their profile and \(L\) for the logarithm of the quantile of \(\xi\) at tail probability \(2r\), we obtain
\[
H(r)=\frac A L+\frac{\kappa A}{L^2}+o(L^{-2}),
\qquad 2r=\frac A{e^L L^2}.
\]
The index resolves the gap once \(L\ge A/\Delta+\kappa+1\), which gives the upper coefficient. For the converse, let \(\lambda_\Delta\) maximize \(f(\lambda)=\mathbb E_{P_\Delta}[\log(1-\lambda X)]\) over \(0<\lambda<1/x_\Delta\), so that \(d_\Delta=f(\lambda_\Delta)\). Reweighting \(P_\Delta\) by a density proportional to \((1-\lambda x)^{-1}\), with \(\lambda\) slightly above \(\lambda_\Delta\), gives a law in the same visible class with positive mean and KL divergence from \(P_\Delta\) close to \(d_\Delta\). At the maximizer, \(\Lambda_\Delta=\log(1/\lambda_\Delta)=A/\Delta+1+O(\Delta)\) and \(d_\Delta=A\lambda_\Delta/\Lambda_\Delta^2(1+o(1))\). The value \(d_\Delta\) coincides with the dual of \citet{honda2015} at level zero. The learner's quantile \(A/\Delta+\kappa+1\) and the converse's \(A/\Delta+1\) differ by a constant, which exponentiates to the gap above. \Cref{cor:continuum} extends the comparison to a continuum of tails.

\section{Consequences and conclusion}\label{sec:conclusion}

\begin{proposition}[Worst case and unknown visibility]\label{prop:consequences}
\textup{(i)} If the first bound below holds for every suboptimal arm, then so does the second:
\[
\mathbb E[N_i(T)]\le\log T\Bigl[C_0+\Bigl(\frac{C_1\sigma}{\Delta_i}\Bigr)^{1/\gamma_\varepsilon}\Bigr],\quad
R_T\le2C_1\sigma\,T^{1-\gamma_\varepsilon}(K\log T)^{\gamma_\varepsilon}+C_0\log T\sum_{i\ne i^\star}\Delta_i.
\]
\textup{(ii)} Let \(\Phi:[1,\infty)\to[1,\infty)\) be non-decreasing and divergent. Restart VT-UCB over epochs of lengths \(3^{2^j}\), \(j\ge0\), with the inputs \((\Phi(t),1/(2\Phi(t)))\) of the starting time \(t\) of each epoch. If the optimal law lies in \(\bigcup_{a\ge0,\,q>0}\mathrm{VCV}(a,q)\), then for all \(T\ge3\),
\[
R_T\le4\,\mathcal B_T\bigl(40+100\,\Phi(T),\,8000\,\Phi(T);1\bigr)+O(1),
\]
where the \(O(1)\) term depends on the instance and on \(\Phi\).

\textup{(iii)} No strategy has \(R_T=O(\log T)\) on every two-armed instance whose optimal law lies in \(\bigcup_{a\ge0,\,q>0}\mathrm{VCV}(a,q)\), even with instance-dependent constants.
\end{proposition}

The price of not knowing the visibility level is therefore a factor that can diverge arbitrarily slowly and cannot be bounded. With known inputs, the same restart scheme makes VT-UCB anytime at the cost of a factor four (\Cref{app:contamination}).

\textbf{Examples.} Every integrable law symmetric about its mean belongs to \(\mathrm{VCV}(\sqrt{3/2},1/4)\). On this class VT-UCB uses no moment parameter and charges a Gaussian competitor the known-parameter rate \(\sigma^2\log T/\Delta^2\) up to the proof constant, whatever the reward origin and the tail of the optimal arm. Translated Pareto laws with tail index in \([3/2,2)\) lie in \(\mathrm{VCV}(34\sqrt2,1/2313)\) (\Cref{app:examples}).

\textbf{Conclusion.} Visibility identifies, law by law, when a heavy-tailed bandit learner can do without moment parameters: hidden upside must stay within the statistical resolution of the rewards already observed. Three questions remain open. Can the universal factor in the scale constant be made independent of the exponent? What is the sharp information constant on general visible laws? Do the lower bounds of \Cref{thm:instance} extend to every competitor?

\bibliography{references}

@inproceedings{agrawal2021,
  title = 	 {Regret Minimization in Heavy-Tailed Bandits},
  author =       {Agrawal, Shubhada and Juneja, Sandeep K. and Koolen, Wouter M.},
  booktitle = 	 {Proceedings of Thirty Fourth Conference on Learning Theory},
  pages = 	 {26--62},
  year = 	 {2021},
  editor = 	 {Belkin, Mikhail and Kpotufe, Samory},
  volume = 	 {134},
  series = 	 {Proceedings of Machine Learning Research},
  month = 	 {15--19 Aug},
  publisher =    {PMLR},
  url = 	 {https://proceedings.mlr.press/v134/agrawal21a.html}
}

@inproceedings{ashutosh2021,
  title = 	 { Bandit algorithms: Letting go of logarithmic regret for statistical robustness },
  author =       {Ashutosh, Kumar and Nair, Jayakrishnan and Kagrecha, Anmol and Jagannathan, Krishna},
  booktitle = 	 {Proceedings of The 24th International Conference on Artificial Intelligence and Statistics},
  pages = 	 {622--630},
  year = 	 {2021},
  editor = 	 {Banerjee, Arindam and Fukumizu, Kenji},
  volume = 	 {130},
  series = 	 {Proceedings of Machine Learning Research},
  month = 	 {13--15 Apr},
  publisher =    {PMLR},
  url = 	 {https://proceedings.mlr.press/v130/ashutosh21a.html}
}

@article{audibert2009,
  title = {{Exploration--exploitation tradeoff using variance estimates in multi-armed bandits}},
  author = {Audibert, Jean-Yves and Munos, R{\'e}mi and Szepesv{\'a}ri, Csaba},
  year = {2009},
  journal = {Theoretical Computer Science},
  volume = {410},
  number = {19},
  pages = {1876--1902},
  doi = {10.1016/j.tcs.2009.01.016},
  url = {https://doi.org/10.1016/j.tcs.2009.01.016},
}

@book{bingham1987,
  title = {{Regular Variation}},
  author = {Bingham, N. H. and Goldie, C. M. and Teugels, J. L.},
  year = {1987},
  publisher = {Cambridge University Press},
  doi = {10.1017/cbo9780511721434},
  url = {https://doi.org/10.1017/cbo9780511721434},
}

@book{boucheron2013,
  title = {{Concentration Inequalities: A Nonasymptotic Theory of Independence}},
  author = {Boucheron, St{\'e}phane and Lugosi, G{\'a}bor and Massart, Pascal},
  year = {2013},
  publisher = {Oxford University Press},
  doi = {10.1093/acprof:oso/9780199535255.001.0001},
  url = {https://academic.oup.com/book/26549},
}

@article{bubeck2013,
  title = {{Bandits with Heavy Tail}},
  author = {Bubeck, S{\'e}bastien and Cesa-Bianchi, Nicol{\`o} and Lugosi, G{\'a}bor},
  journal = {IEEE Transactions on Information Theory},
  volume = {59},
  number = {11},
  pages = {7711--7717},
  year = {2013},
  doi = {10.1109/TIT.2013.2277869},
  url = {https://doi.org/10.1109/TIT.2013.2277869},
}

@inproceedings{chen2025,
  title = {{uniINF: Best-of-Both-Worlds Algorithm for Parameter-Free Heavy-Tailed MABs}},
  author = {Chen, Yu and Huang, Jiatai and Dai, Yan and Huang, Longbo},
  year = {2025},
  booktitle = {International Conference on Learning Representations},
  url = {https://proceedings.iclr.cc/paper_files/paper/2025/hash/b2c4b7d34b3d96b9dc12f7bce424b7ae-Abstract-Conference.html},
}

@inproceedings{genalti2024,
  title = 	 {$(\varepsilon, u)$-Adaptive Regret Minimization in Heavy-Tailed Bandits},
  author =       {Genalti, Gianmarco and Marsigli, Lupo and Gatti, Nicola and Metelli, Alberto Maria},
  booktitle = 	 {Proceedings of Thirty Seventh Conference on Learning Theory},
  pages = 	 {1882--1915},
  year = 	 {2024},
  editor = 	 {Agrawal, Shipra and Roth, Aaron},
  volume = 	 {247},
  series = 	 {Proceedings of Machine Learning Research},
  month = 	 {30 Jun--03 Jul},
  publisher =    {PMLR},
  url = 	 {https://proceedings.mlr.press/v247/genalti24a.html}
}

@inproceedings{genalti2025,
  title = 	 {Open Problem: Regret Minimization in Heavy-Tailed Bandits with Unknown Distributional Parameters},
  author =       {Genalti, Gianmarco and Metelli, Alberto Maria},
  booktitle = 	 {Proceedings of Thirty Eighth Conference on Learning Theory},
  pages = 	 {1--5},
  year = 	 {2025},
  editor = 	 {Haghtalab, Nika and Moitra, Ankur},
  volume = 	 {291},
  series = 	 {Proceedings of Machine Learning Research},
  month = 	 {30 Jun--04 Jul},
  publisher =    {PMLR},
  url = 	 {https://proceedings.mlr.press/v291/genalti25a.html}
}

@misc{genalti2026,
  title = {{Parameter-Free Heavy-Tailed Bandits}},
  author = {Genalti, Gianmarco and Metelli, Alberto Maria},
  year = {2026},
  eprint = {2607.29460},
  archivePrefix = {arXiv},
  primaryClass = {cs.LG},
  note = {Preprint, version 1, July 31, 2026},
  doi = {10.48550/arXiv.2607.29460},
  url = {https://arxiv.org/abs/2607.29460},
}

@article{hadiji2023,
  title = {{Adaptation to the Range in K-Armed Bandits}},
  author = {H{\'e}di Hadiji and Gilles Stoltz},
  year = {2023},
  journal = {Journal of Machine Learning Research},
  volume = {24},
  number = {13},
  pages = {1--33},
  url = {https://www.jmlr.org/papers/v24/21-0148.html},
}

@inproceedings{huang2022,
  title = 	 {Adaptive Best-of-Both-Worlds Algorithm for Heavy-Tailed Multi-Armed Bandits},
  author =       {Huang, Jiatai and Dai, Yan and Huang, Longbo},
  booktitle = 	 {Proceedings of the 39th International Conference on Machine Learning},
  pages = 	 {9173--9200},
  year = 	 {2022},
  editor = 	 {Chaudhuri, Kamalika and Jegelka, Stefanie and Song, Le and Szepesvari, Csaba and Niu, Gang and Sabato, Sivan},
  volume = 	 {162},
  series = 	 {Proceedings of Machine Learning Research},
  month = 	 {17--23 Jul},
  publisher =    {PMLR},
  url = 	 {https://proceedings.mlr.press/v162/huang22c.html}
}

@article{lai1985,
  title = {{Asymptotically efficient adaptive allocation rules}},
  author = {Lai, T. L. and Robbins, Herbert},
  year = {1985},
  journal = {Advances in Applied Mathematics},
  volume = {6},
  number = {1},
  pages = {4--22},
  doi = {10.1016/0196-8858(85)90002-8},
  url = {https://doi.org/10.1016/0196-8858(85)90002-8},
}

@inproceedings{lattimore2017,
  title = {{A Scale Free Algorithm for Stochastic Bandits with Bounded Kurtosis}},
  author = {Lattimore, Tor},
  year = {2017},
  booktitle = {Advances in Neural Information Processing Systems},
  volume = {30},
  url = {https://papers.nips.cc/paper_files/paper/2017/hash/fed33392d3a48aa149a87a38b875ba4a-Abstract.html},
}

@book{lattimore2020,
  title = {{Bandit Algorithms}},
  author = {Lattimore, Tor and Szepesv{\'a}ri, Csaba},
  year = {2020},
  publisher = {Cambridge University Press},
  doi = {10.1017/9781108571401},
  url = {https://doi.org/10.1017/9781108571401},
}

@inproceedings{lee2023,
  title = 	 {Optimality of {Thompson} Sampling with Noninformative Priors for {Pareto} Bandits},
  author =       {Lee, Jongyeong and Honda, Junya and Chiang, Chao-Kai and Sugiyama, Masashi},
  booktitle = 	 {Proceedings of the 40th International Conference on Machine Learning},
  pages = 	 {18810--18851},
  year = 	 {2023},
  editor = 	 {Krause, Andreas and Brunskill, Emma and Cho, Kyunghyun and Engelhardt, Barbara and Sabato, Sivan and Scarlett, Jonathan},
  volume = 	 {202},
  series = 	 {Proceedings of Machine Learning Research},
  month = 	 {23--29 Jul},
  publisher =    {PMLR},
  url = 	 {https://proceedings.mlr.press/v202/lee23d.html}
}

@inproceedings{maurer2009,
  title = {{Empirical Bernstein Bounds and Sample Variance Penalization}},
  author = {Maurer, Andreas and Pontil, Massimiliano},
  booktitle = {Proceedings of the 22nd Annual Conference on Learning Theory},
  year = {2009},
  url = {https://www.cs.mcgill.ca/~colt2009/papers/012.pdf},
}

@misc{oliveira2025,
  title = {{Finite-sample Properties of the Trimmed Mean}},
  author = {Oliveira, Roberto I. and Orenstein, Paulo and Rico, Zoraida F.},
  year = {2025},
  eprint = {2501.03694},
  archivePrefix = {arXiv},
  primaryClass = {math.ST},
  note = {Preprint},
  doi = {10.48550/arXiv.2501.03694},
  url = {https://arxiv.org/abs/2501.03694},
}

@inproceedings{tamas2025,
  title = 	 {Data-Driven Upper Confidence Bounds with Near-Optimal Regret for Heavy-Tailed Bandits},
  author =       {Tam{\'a}s, Ambrus and Szentp{\'e}teri, Szabolcs and Cs{\'a}ji, Bal{\'a}zs},
  booktitle = 	 {Proceedings of The 28th International Conference on Artificial Intelligence and Statistics},
  pages = 	 {2116--2124},
  year = 	 {2025},
  editor = 	 {Li, Yingzhen and Mandt, Stephan and Agrawal, Shipra and Khan, Emtiyaz},
  volume = 	 {258},
  series = 	 {Proceedings of Machine Learning Research},
  month = 	 {03--05 May},
  publisher =    {PMLR},
  url = 	 {https://proceedings.mlr.press/v258/tamas25a.html}
}

@book{tsybakov2009,
  title = {{Introduction to Nonparametric Estimation}},
  author = {Tsybakov, Alexandre B.},
  year = {2009},
  publisher = {Springer New York},
  doi = {10.1007/b13794},
  url = {https://doi.org/10.1007/b13794},
}

@article{bahadur1956,
  title = {{The Nonexistence of Certain Statistical Procedures in Nonparametric Problems}},
  author = {Bahadur, R. R. and Savage, Leonard J.},
  journal = {The Annals of Mathematical Statistics},
  volume = {27},
  number = {4},
  pages = {1115--1122},
  year = {1956},
  doi = {10.1214/aoms/1177728077},
  url = {https://doi.org/10.1214/aoms/1177728077},
}

@article{catoni2012,
  title = {{Challenging the empirical mean and empirical variance: A deviation study}},
  author = {Catoni, Olivier},
  year = {2012},
  journal = {Annales de l'Institut Henri Poincar{\'e}, Probabilit{\'e}s et Statistiques},
  volume = {48},
  number = {4},
  pages = {1148--1185},
  doi = {10.1214/11-aihp454},
  url = {https://doi.org/10.1214/11-aihp454},
}

@article{devroye2016,
  title = {{Sub-Gaussian mean estimators}},
  author = {Devroye, Luc and Lerasle, Matthieu and Lugosi, G{\'a}bor and Oliveira, Roberto I.},
  year = {2016},
  journal = {The Annals of Statistics},
  volume = {44},
  number = {6},
  pages = {2695--2725},
  doi = {10.1214/16-aos1440},
  url = {https://doi.org/10.1214/16-aos1440},
}

@article{lugosi2019,
  title = {{Mean Estimation and Regression Under Heavy-Tailed Distributions: A Survey}},
  author = {Lugosi, G{\'a}bor and Mendelson, Shahar},
  year = {2019},
  journal = {Foundations of Computational Mathematics},
  volume = {19},
  number = {5},
  pages = {1145--1190},
  doi = {10.1007/s10208-019-09427-x},
  url = {https://doi.org/10.1007/s10208-019-09427-x},
}

@article{kaufmann2016,
  title = {{On the Complexity of Best-Arm Identification in Multi-Armed Bandit Models}},
  author = {Emilie Kaufmann and Olivier Capp{\'e} and Aur{\'e}lien Garivier},
  year = {2016},
  journal = {Journal of Machine Learning Research},
  volume = {17},
  number = {1},
  pages = {1--42},
  url = {https://jmlr.org/papers/v17/kaufman16a.html},
}

@inproceedings{cheng2024,
  title = {{Taming Heavy-Tailed Losses in Adversarial Bandits and the Best-of-Both-Worlds Setting}},
  author = {Cheng, Duo and Zhou, Xingyu and Ji, Bo},
  year = {2024},
  booktitle = {Advances in Neural Information Processing Systems},
  volume = {37},
  pages = {48083--48129},
  doi = {10.52202/079017-1524},
  url = {https://papers.nips.cc/paper_files/paper/2024/hash/55f5aa8d997a8071d0651e0009ec49a2-Abstract-Conference.html},
}

@inproceedings{cherapanamjeri2025,
  title = 	 {Heavy-tailed Estimation is Easier than Adversarial Contamination},
  author =       {Cherapanamjeri, Yeshwanth and Lee, Daniel},
  booktitle = 	 {Proceedings of Thirty Eighth Conference on Learning Theory},
  pages = 	 {1154--1184},
  year = 	 {2025},
  editor = 	 {Haghtalab, Nika and Moitra, Ankur},
  volume = 	 {291},
  series = 	 {Proceedings of Machine Learning Research},
  month = 	 {30 Jun--04 Jul},
  publisher =    {PMLR},
  url = 	 {https://proceedings.mlr.press/v291/cherapanamjeri25a.html}
}

@inproceedings{hammouda2026,
  title = {{Sub-Gaussian Confidence Intervals for Heavy-Tailed Data: Characterizing the Limits of Inference}},
  author = {Hammouda, Ilyes and Minsker, Stanislav and Ndaoud, Mohamed},
  booktitle = {Advances in Neural Information Processing Systems},
  year = {2026},
  eprint = {2610.15799},
  archivePrefix = {arXiv},
  primaryClass = {math.ST},
  url = {https://arxiv.org/abs/2610.15799}
}

@inproceedings{honda2010,
  title = {{An Asymptotically Optimal Bandit Algorithm for Bounded Support Models}},
  author = {Honda, Junya and Takemura, Akimichi},
  booktitle = {Proceedings of COLT 2010},
  pages = {67--79},
  year = {2010},
}

@article{honda2015,
  title = {{Non-Asymptotic Analysis of a New Bandit Algorithm for Semi-Bounded Rewards}},
  author = {Honda, Junya and Takemura, Akimichi},
  journal = {Journal of Machine Learning Research},
  volume = {16},
  number = {113},
  pages = {3721--3756},
  year = {2015},
  url = {https://jmlr.org/papers/v16/honda15a.html},
}

@misc{gupta2026,
  title = {{Reliable Scale-Free Bandits via Empirical Likelihood}},
  author = {Gupta, Pratham and Sethi, Alhad and Agrawal, Shubhada},
  year = {2026},
  eprint = {2610.18055},
  archivePrefix = {arXiv},
  primaryClass = {stat.ML},
  note = {Preprint, version 1, October 13, 2026},
  url = {https://arxiv.org/abs/2610.18055}
}

\appendix
\section{Threshold forms of visibility}\label{app:converse}

Trimming keeps the observations within distance \(M\) of a center \(z\). For \(Y=X-z\), define the trimmed probability, the retained second moment, the mean deficit and the width scale:
\[
\begin{aligned}
p_z(M)&=\mathbb P[|Y|>M],&V_z(M)&=\mathbb E[Y^2\mathbf1_{\{|Y|\le M\}}],\\
b_z(M)&=\mathbb E[(X-\mu)\mathbf1_{\{|Y|>M\}}],&
s_z(M)&=\sqrt{p_z(M)V_z(M)}+Mp_z(M).
\end{aligned}
\]
We say that \(X-z\) has threshold visibility, written \(X-z\in\mathrm{VT}(c,q)\), when \(b_z(M)\le cs_z(M)\) for all \(M\ge0\) with \(p_z(M)\le q\). At \(z=0\), this is the raw-moment visibility condition, and we then drop the subscript \(z\).

\begin{proposition}[Threshold forms]\label{prop:threshold}
For a class \(\mathcal D\) of integrable laws the following are equivalent: \textup{(a)} \(\mathcal D\subseteq\mathrm{VCV}(a,q)\) for some finite \(a\) and positive \(q\); \textup{(b)} uniform constants \((c,q)\) make \(X-\mu\in\mathrm{VT}(c,q)\) for every law in \(\mathcal D\); \textup{(c)} uniform constants \((c,q)\) make \(X-z\in\mathrm{VT}(c,q)\) for every law in \(\mathcal D\) and every \(z\in\mathbb R\). Moreover \(\mathrm{VCV}(a,q)\) implies \(\mathrm{VT}(a,q)\) at every center, and \(\mathrm{VT}(c,q)\) at the mean implies \(\mathrm{VCV}(\sqrt2\,c',\min\{q/2,1/4,(2c'^2+1)^{-1}\})\) with \(c'=6c+4\).
\end{proposition}

Throughout, conditional quantities are used only for sets of positive probability. The omitted centered mean can be negative, in which case every visibility inequality is automatic. We therefore square inequalities only after restricting to a positive omitted mean. All retained second moments in this appendix are finite at finite truncation thresholds.

For an integrable law, write \(p=\mathbb P[E^c]\), \(b_E=\mathbb E[(X-\mu)\mathbf1_{E^c}]\), and let \(v=\mathbb E[X^2\mathbf1_E]\) be the raw analogue of \(v_E\). Raw core visibility \((a,q)\) means \(b_E\le a\sqrt{pv}\) whenever \(0<p\le q\) and \(v<\infty\). It immediately implies threshold visibility by taking \(E=\{|X|\le M\}\).

Conversely, suppose VT\((c,q)\) holds. For \(0<p\le\min\{q/2,1/4\}\), first assume \(v>0\) and \(b_E>0\). Set \(t=\sqrt{v/p}\). Markov's inequality within \(E\) gives
\[
p(t)\le p+v/t^2=2p,\qquad V(t)\le v+pt^2=2v,\qquad b(t)\le4c\sqrt{pv}.
\]
The difference between the omitted integrals is supported on \(E^c\cap\{|X|\le t\}\) and \(E\cap\{|X|>t\}\). Since \(|p(t)-p|\le p\),
\[
b_E\le b(t)+pt+v/t+p|\mu|.
\]
Using \((1-p)\mu=\mathbb E[X\mathbf1_E]+b_E\) and Cauchy-Schwarz yields
\[
\frac{1-2p}{1-p}b_E\le(4c+2)\sqrt{pv}+p\sqrt{v/(1-p)}
\le\frac{1-2p}{1-p}(6c+4)\sqrt{pv}.
\]
If \(v=0\), then \(X=0\) on \(E\). For every \(t>0\), \(p(t)\le p\) and \(V(t)\le pt^2\), so \(b(t)\le2cpt\). Dominated convergence gives \(\mu\mathbb P[X=0]\le0\), hence \(b_E=(1-p)\mu\le0\). This proves the raw equivalence, including the case \(v=0\). The threshold \(t\) is chosen from the retained second moment of \(E\), which removes the term \(Mp(M)\) from the final bound. The reverse implication requires no change of constants.

For the centered equivalence, put \(v_E=(1-p)\operatorname{Var}[X\mid E]\). The identity
\begin{equation}\label{eq:retained-variance}
\mathbb E[(X-z)^2\mathbf1_E]=v_E+(1-p)(\mu_E-z)^2
\end{equation}
shows that VCV\((a,q)\) implies VT\((a,q)\) for \(X-z\) at every \(z\). Conversely, uniform threshold visibility over all \(z\) implies VCV\((6c+4,\min\{q/2,1/4\})\) by applying the raw equivalence with \(z=\mu_E\). If visibility is known only about \(\mu\), put \(c'=6c+4\). For \(b_E>0\), the raw equivalence and \eqref{eq:retained-variance} give
\[
b_E^2\le c'^2p\left(v_E+\frac{b_E^2}{1-p}\right).
\]
For \(p\le(2c'^2+1)^{-1}\), moving the \(b_E^2\) term to the left side gives \(b_E^2\le2c'^2pv_E\). This proves VCV\((\sqrt2\,c',\min\{q/2,1/4,(2c'^2+1)^{-1}\})\).

\section{Converse, exchange and class frontier}\label{app:exchange}

\begin{lemma}[Information requirement]\label{lem:information}
Let two instances differ in one arm only, with laws \(\pi\) and \(\nu\) for that arm. Suppose the arm is suboptimal in the first instance and uniquely optimal in the second, and the strategy is uniformly efficient on both. Then, with \(\mathbb E_0\) the expectation in the first instance,
\[
\liminf_{T\to\infty}\frac{\mathbb E_0[N_\pi(T)]}{\log T}\ge\frac1{\operatorname{KL}(\pi\Vert\nu)}.
\]
\end{lemma}
\begin{proof}
Assume the divergence is finite and let \(F_T=\{N_\pi(T)<T/2\}\). In the first instance the arm has a positive gap, so \(\mathbb P_0[F_T^c]\le2\mathbb E_0[N_\pi(T)]/T\to0\). In the second instance every other arm has a gap bounded below, so \(\mathbb P_1[F_T]\le2\mathbb E_1[T-N_\pi(T)]/T=o(T^{-1+\delta})\) for every \(\delta>0\). The divergence decomposition and data processing for the binary relative entropy \(\operatorname{kl}\) \citep{kaufmann2016, tsybakov2009} give
\[
\mathbb E_0[N_\pi(T)]\operatorname{KL}(\pi\Vert\nu)\ge\operatorname{kl}(\mathbb P_0[F_T],\mathbb P_1[F_T])
\ge\mathbb P_0[F_T]\log\frac1{\mathbb P_1[F_T]}-\log2.
\]
Divide by \(\log T\), take the lower limit, and let \(\delta\downarrow0\).
\end{proof}

\begin{theorem}[Law-by-law converse at every exponent]\label{thm:converse}
Fix \(0<\varepsilon\le1\) and let a strategy have budget \((C_0,C_1)\) at exponent \(\varepsilon\). Let \(\nu\) be an integrable law and \(E\) a set with \(0<p\le\min\{1/2,1/(8C_0)\}\), \(\sigma_{E,\varepsilon}<\infty\) and \(b_E>4C_1p^{\gamma_\varepsilon}\sigma_{E,\varepsilon}\). With \(\Delta\), \(\theta\) and \(d\) as in \Cref{thm:frontier}, put
\[
\eta=2d\Bigl[C_0+(1-p)\bigl(2C_1\sigma_{E,\varepsilon}/b_E\bigr)^{1/\gamma_\varepsilon}\Bigr].
\]
Then \(\eta<1\) and \(R_T(\delta_\theta,\nu)\ge\frac\Delta4T^{1-\eta}\) whenever \(2\eta\log T\le dT\).
\end{theorem}
\begin{proof}
Follow the proof of \Cref{thm:frontier}(ii). The conditional law \(\pi=\nu(\cdot\mid E)\) has \(\sigma_\varepsilon(\pi)=\sigma_{E,\varepsilon}(1-p)^{-1/(1+\varepsilon)}\), so
\[
\Bigl(\frac{C_1\sigma_\varepsilon(\pi)}\Delta\Bigr)^{1/\gamma_\varepsilon}
=(1-p)\Bigl(\frac{2C_1\sigma_{E,\varepsilon}}{b_E}\Bigr)^{1/\gamma_\varepsilon},
\]
and the budget gives \(\mathbb E_0[N_2(T)]\le\eta\log T/(2d)\). The likelihood argument is unchanged. For the exponent, the violation gives \((2C_1\sigma_{E,\varepsilon}/b_E)^{1/\gamma_\varepsilon}<2^{-1/\gamma_\varepsilon}/p\), so the second term of \(\eta\) is below \(2^{1-1/\gamma_\varepsilon}(1-p)d/p\le1/2\), because \(1/\gamma_\varepsilon\ge2\). The first term is at most \(1/2\) as before.
\end{proof}

\begin{proof}[Proof of \Cref{thm:frontier}\textup{(iii)}]
\emph{Allowance.} For \(q<p<1\) let \(\nu=(1-p)\delta_0+p\delta_1\). Its upper atom has mass above \(q\), and omitting the lower atom gives \(b_E<0\), so \(\nu\in\mathrm{VCV}(0,q)\subseteq\mathrm{VCV}(a,q)\). Compare \((\delta_{p/2},\delta_0)\) with \((\delta_{p/2},\nu)\). The competitor \(\delta_0\) has zero variance, so the budget gives \(\mathbb E_0[N_2(T)]\le C_0\log T\), while \(\operatorname{KL}(\delta_0\Vert\nu)=\log\frac1{1-p}\). \Cref{lem:information} gives \(C_0\log\frac1{1-p}\ge1\), and \(p\downarrow q\) gives the first bound. The second follows from \(\log\frac1{1-q}\le q/(1-q)\).

\emph{Scale.} Let \(\pi=\frac34\delta_0+\frac14\delta_{-1}\), with mean \(-1/4\) and variance \(\sigma^2=3/16\). For small \(\Delta>0\) take the first instance \((\delta_{-1/4+\Delta},\pi)\) and put
\[
\iota=\frac{2(1+\sqrt\Delta)\Delta}{a^2+2},\qquad p=\frac{a^2(1+\sqrt\Delta)^2\Delta^2}{\sigma^2(a^2+2)^2},\qquad
\pi_\iota=\bigl(\tfrac34+\iota\bigr)\delta_0+\bigl(\tfrac14-\iota\bigr)\delta_{-1}.
\]
The tilted law \(\pi_\iota\) has mean \(\mu_\iota=-1/4+\iota\) and standard deviation \(\sigma_\iota=\sigma+O(\Delta)\). For \(a>0\) let
\[
x=\mu_\iota+\frac{a\sigma_\iota}{\sqrt{p(1-p)}},\qquad\nu_\Delta=(1-p)\pi_\iota+p\,\delta_x,
\]
and for \(a=0\) let \(\nu_\Delta=\pi_\iota\). For small \(\Delta\) the atom at \(0\) has mass above \(1/2\), \(x>0\), and \(\nu_\Delta\) has negative mean. Omitting \(\{-1\}\) gives \(b_E<0\). Omitting \(\{-1,x\}\) retains the atom at \(0\) alone, where \(v_E=0\) and \(b_E<0\). Omitting \(\{x\}\) gives \(v_E=(1-p)\sigma_\iota^2\) and \(b_E=p(1-p)(x-\mu_\iota)=a\sqrt{pv_E}\). Hence \(\nu_\Delta\in\mathrm{VCV}(a,q)\). Its mean exceeds that of \(\pi\) by
\[
\iota+a\sigma_\iota\sqrt{\frac p{1-p}}=(1+\sqrt\Delta)\Delta\,(1+O(\Delta))>\Delta
\]
for small \(\Delta\), so \(\nu_\Delta\) is optimal in the second instance \((\delta_{-1/4+\Delta},\nu_\Delta)\). Since \(x\notin\{-1,0\}\),
\[
\operatorname{KL}(\pi\Vert\nu_\Delta)=\log\frac1{1-p}+\operatorname{KL}(\pi\Vert\pi_\iota)=p+\frac{\iota^2}{2\sigma^2}+O(\Delta^3)=\frac{(1+\sqrt\Delta)^2\Delta^2}{\sigma^2(a^2+2)}+O(\Delta^3).
\]
The pair \((\iota,p)\) minimizes \(p+\iota^2/(2\sigma^2)\) for the mean gain \(\iota+a\sigma\sqrt p\). The budget and \Cref{lem:information} give \((C_0+C_1^2\sigma^2/\Delta^2)\operatorname{KL}(\pi\Vert\nu_\Delta)\ge1\), and \(\Delta\downarrow0\) gives \(C_1^2\ge a^2+2\).
\end{proof}

The rare atom alone gives a bound at every exponent. For \(0<\varepsilon\le1\), \(a>0\) and small \(p\), let \(\pi\) be as above and \(\nu=(1-p)\pi+p\,\delta_x\) with \(x=-1/4+a\,\sigma_\varepsilon(\pi)\,p^{\gamma_\varepsilon-1}(1-p)^{-\gamma_\varepsilon}\). Omitting \(\{x\}\) gives \(\sigma_{E,\varepsilon}=(1-p)^{1/(1+\varepsilon)}\sigma_\varepsilon(\pi)\) and \(b_E=a\,p^{\gamma_\varepsilon}\sigma_{E,\varepsilon}\), and the other omissions have \(b_E<0\), so \(\nu\in\mathcal V_\varepsilon(a,q)\). The mean gain is \(2\Delta_p\) with \(\Delta_p=\frac12a\,\sigma_\varepsilon(\pi)\,p^{\gamma_\varepsilon}(1-p)^{-\gamma_\varepsilon}\). Comparing \((\delta_{-1/4+\Delta_p},\pi)\) with \((\delta_{-1/4+\Delta_p},\nu)\), a budget \((C_0,C_1)\) at exponent \(\varepsilon\) and \Cref{lem:information} give
\[
\Bigl[C_0+\Bigl(\frac{2C_1}a\Bigr)^{1/\gamma_\varepsilon}\frac{1-p}p\Bigr]\log\frac1{1-p}\ge1,
\]
and \(p\downarrow0\) gives \(C_1\ge a/2\) for every strategy that is uniformly efficient on \(\mathcal V_\varepsilon(a,q)\).

\begin{proof}[Proof of \Cref{cor:class}]
Sufficiency is \Cref{thm:integrability}. For necessity, point masses belong to \(\mathcal D\), so the strategy has budget \((C_0,C_1)\) at exponent \(\bar\varepsilon\). If some \(\nu\in\mathcal D\) lay outside \(\mathcal V_{\bar\varepsilon}(4C_1,\min\{1/2,1/(8C_0)\})\), \Cref{thm:converse} would give \(R_T(\delta_\theta,\nu)\ge\frac\Delta4T^{1-\eta}\) with \(\eta<1\). The law \(\nu\) is optimal in \((\delta_\theta,\nu)\) and the competitor has zero scale, so the assumed bound is \(C_0\Delta\log T\), a contradiction for large \(T\).
\end{proof}

The point masses can be replaced by a translation-closed class \(\mathcal D\) that contains one law with a finite central moment of some order \(1+\varepsilon\), \(\varepsilon\le\bar\varepsilon\). The translate of that law with mean \(\theta\) then plays the role of \(\delta_\theta\) in both instances, and the likelihood argument is unchanged.

\section{Quantile trimming with atoms}\label{app:trimming}

\begin{lemma}[Quantile trimming]\label{lem:trimming}
Let \(X_1,\ldots,X_n\) be independent observations from an integrable law \(\nu\), let \(\ell\ge1\) be an integer with \(n\ge4\ell\), and put \(r=\ell/n\). Discard the \(\ell\) largest absolute observations, breaking ties by index. Define \(\widehat M,\widehat\mu,\widehat V,\widehat w\) as in the algorithm with pilot center zero, and set
\[
M_- =\inf\{M\ge0:p(M)\le2r\},\qquad M_+=\inf\{M\ge0:p(M)\le r/2\}.
\]
Write \(p_\pm=p(M_\pm)\), \(V_\pm=V(M_\pm)\), and \(\mu_-=\mathbb E[X\mid |X|\le M_-]\). The following hold jointly with probability at least \(1-7e^{-\ell/6}\):
\begin{equation}\label{eq:quantile-trimming}
M_-\le\widehat M\le M_+,\quad
|\widehat\mu-\mu_-|\le3\widehat w,\quad
s(M_-)\le\tfrac32\widehat w,\quad
\widehat w\le W_\nu(r).
\end{equation}
If \((\mathbb E[|X|^{1+\varepsilon}])^{1/(1+\varepsilon)}\le\sigma\) for some \(0<\varepsilon\le1\), also \(\widehat w\le25\,\sigma\,r^{\gamma_\varepsilon}\).
\end{lemma}
\begin{proof}

Right continuity gives \(p(M_-)\le2r\) and \(p(M_+)\le r/2\). At a positive quantile, the inclusive tail probability is at least its defining level. Binomial Chernoff bounds consequently give the events
\[
\begin{aligned}
\mathcal G_1&=\{\#\{|X_j|>M_+\}\le\ell\},\\
\mathcal G_2&=\{\#\{|X_j|\ge M_-\}\ge\ell+1\},\\
\mathcal G_3&=\{\#\{|X_j|>M_-\}\le4\ell\},
\end{aligned}
\]
with failure probabilities at most \(e^{-\ell/6},e^{-\ell/4},e^{-2\ell/3}\). Take \(\mathcal G_2\) certain when \(M_-=0\). These choices distinguish atoms on a threshold from observations strictly beyond it. On \(\mathcal G_1\), at most the discarded number of observations exceed \(M_+\). On \(\mathcal G_2\), at least one of the observations at or above \(M_-\) remains. Thus their intersection gives the claimed empirical threshold bracket regardless of the tie order. The defining population quantiles are finite by integrability. If a quantile is positive, approaching it from below proves the inclusive-tail assertion by continuity of probabilities.

Put \(\bar X_n=n^{-1}\sum_jX_j\mathbf1_{\{|X_j|\le M_-\}}\) and \(\bar x_-=(1-p_- )\mu_-\).

Bernstein's inequality \citep{boucheron2013} gives
\[
|\bar X_n-\bar x_-|\le\sqrt{2V_-r}+2M_-r/3
\]
except with probability \(2e^{-\ell}\). On \(\mathcal G_2\cap\mathcal G_3\), the retained sum differs from \(n\bar X_n\) by at most \(5\ell\widehat M\). At most \(4\ell\) retained observations lie above \(M_-\), and at most \(\ell\) discarded observations lie at or below it. Using
\[
\frac n{n-\ell}\le\frac43,\quad |\bar x_-|\le M_-,\quad
\left|\frac n{n-\ell}-\frac1{1-p_-}\right|\le4r
\]
in the exact decomposition
\[
\widehat\mu-\mu_-=\frac{\sum_{j\in S}X_j-n\bar X_n}{n-\ell}
+\frac n{n-\ell}(\bar X_n-\bar x_-)
+\bar x_-\left(\frac n{n-\ell}-\frac1{1-p_-}\right)
\]
gives
\begin{equation}\label{eq:trimmed-mean-error}
|\widehat\mu-\mu_-|\le\tfrac43\sqrt{2V_-r}+12r\widehat M.
\end{equation}
For \(\overline{X^2}_-=n^{-1}\sum_jX_j^2\mathbf1_{\{|X_j|\le M_-\}}\), another Bernstein event has failure probability \(e^{-\ell}\) and gives
\[
\overline{X^2}_-\ge V_- -M_-\sqrt{2V_-r}-2M_-^2r/3,
\qquad \widehat V\ge\overline{X^2}_- -4M_-^2r/3.
\]
If \(V_-\ge36M_-^2r\), these losses are smaller than \(V_-/2\). Otherwise \(\sqrt{2V_-r}\le\sqrt{72}M_-r\). Hence
\[
\sqrt{2V_-r}\le\sqrt2\sqrt{2\widehat Vr}+8.5r\widehat M.
\]
Writing \(\hat h=\sqrt{2\widehat Vr}\) and \(\hat t=r\widehat M\), the right side of \eqref{eq:trimmed-mean-error} is at most \(4\sqrt2\hat h/3+(70/3)\hat t\le3\hat h+27\hat t\). Also
\[
s(M_-)\le\sqrt{2V_-r}+2M_-r\le\sqrt2\hat h+10.5\hat t\le1.5(\hat h+9\hat t).
\]
These inequalities prove the middle two conclusions of \eqref{eq:quantile-trimming}.

Finally, Bernstein and \(xy\le(x^2+y^2)/2\) give, except with probability \(e^{-\ell}\),
\[
\overline{X^2}_+:=n^{-1}\sum_jX_j^2\mathbf1_{\{|X_j|\le M_+\}}
\le2V_++\tfrac76M_+^2r.
\]
On \(\mathcal G_1\), \(\widehat V\le4\overline{X^2}_+/3\), so
\[
\widehat w\le\frac4{\sqrt3}\sqrt{V_+r}
+\left(9+\frac{\sqrt{28}}3\right)M_+r
\le3\sqrt{V_+r}+11M_+r=W_\nu(r).
\]
For the moment conclusion, the inclusive quantile bound and Markov's inequality give, with \(w_{\mathrm m}=\sigma r^{\gamma_\varepsilon}\),
\[
M_+r\le2w_{\mathrm m},\qquad V_+r\le2(w_{\mathrm m})^2.
\]
The preceding sharper bound is at most \(25w_{\mathrm m}\). If \(M_+=0\), the retained observations vanish on \(\mathcal G_1\). Summing the six failure probabilities proves the lemma.
\end{proof}

For \(\bar\varepsilon\in(0,1]\) define the retained moment of order \(1+\bar\varepsilon\), its population counterparts and the corresponding width,
\[
\widehat V^{(\bar\varepsilon)}=\frac{\sum_{j\in S}|X_j|^{1+\bar\varepsilon}}{n-\ell},\ \
V^{(\bar\varepsilon)}_\pm=\mathbb E\bigl[|X|^{1+\bar\varepsilon}\mathbf1_{\{|X|\le M_\pm\}}\bigr],\ \
\widehat w^{(\bar\varepsilon)}=r^{\gamma_{\bar\varepsilon}}\bigl(\widehat V^{(\bar\varepsilon)}\bigr)^{\frac1{1+\bar\varepsilon}}+9r\widehat M.
\]

\begin{lemma}[Trimming with a retained moment]\label{lem:trimming-moment}
In the setting of \Cref{lem:trimming}, fix \(\bar\varepsilon\in(0,1]\). With probability at least \(1-9e^{-\ell/6}\), \eqref{eq:quantile-trimming} holds together with
\begin{equation}\label{eq:moment-trimming}
|\widehat\mu-\mu_-|\le3\sqrt2\,\widehat w^{(\bar\varepsilon)},\qquad
V_-^{(\bar\varepsilon)}\le2\widehat V^{(\bar\varepsilon)}+6r\widehat M^{1+\bar\varepsilon},\qquad
\widehat V^{(\bar\varepsilon)}\le\tfrac83V_+^{(\bar\varepsilon)}+\tfrac{14}9rM_+^{1+\bar\varepsilon}.
\end{equation}
If \((\mathbb E[|X|^{1+\varepsilon}])^{1/(1+\varepsilon)}\le\sigma\) for some \(0<\varepsilon\le\bar\varepsilon\), then on the same event \(\widehat w^{(\bar\varepsilon)}\le25\,\sigma\,r^{\gamma_\varepsilon}\).
\end{lemma}
\begin{proof}
Work on the event of \Cref{lem:trimming}. Each retained observation has \(|X_j|\le\widehat M\), so \(\widehat V\le\widehat M^{1-\bar\varepsilon}\widehat V^{(\bar\varepsilon)}\). With \(\hat h=r^{\gamma_{\bar\varepsilon}}(\widehat V^{(\bar\varepsilon)})^{1/(1+\bar\varepsilon)}\) and \(\hat t=r\widehat M\), the weighted arithmetic--geometric mean inequality gives
\[
\sqrt{2r\widehat V}\le\sqrt2\,\hat h^{(1+\bar\varepsilon)/2}\hat t^{(1-\bar\varepsilon)/2}\le\sqrt2(\hat h+\hat t),
\]
so \(\widehat w\le\sqrt2\hat h+(9+\sqrt2)\hat t\le\sqrt2\,\widehat w^{(\bar\varepsilon)}\), and the first bound follows from \eqref{eq:quantile-trimming}.

The summands \(|X_j|^{1+\bar\varepsilon}\mathbf1_{\{|X_j|\le M_-\}}\) lie in \([0,M_-^{1+\bar\varepsilon}]\) and have second moment at most \(M_-^{1+\bar\varepsilon}V_-^{(\bar\varepsilon)}\). Bernstein's inequality and \(\sqrt{2xy}\le x/2+y\) show that, except with probability \(e^{-\ell}\), their average is at least \(\frac12V_-^{(\bar\varepsilon)}-\frac53M_-^{1+\bar\varepsilon}r\). On \(\mathcal G_2\) at most \(\ell\) of these observations are discarded, which costs at most \(\frac43M_-^{1+\bar\varepsilon}r\). Hence \(\widehat V^{(\bar\varepsilon)}\ge\frac12V_-^{(\bar\varepsilon)}-3M_-^{1+\bar\varepsilon}r\), and \(M_-\le\widehat M\) gives the second bound. In the same way, except with probability \(e^{-\ell}\), the average of \(|X_j|^{1+\bar\varepsilon}\mathbf1_{\{|X_j|\le M_+\}}\) is at most \(2V_+^{(\bar\varepsilon)}+\frac76M_+^{1+\bar\varepsilon}r\), by \(\sqrt{2xy}\le x+y/2\). On \(\mathcal G_1\) all retained observations satisfy \(|X_j|\le M_+\), and \(n/(n-\ell)\le4/3\) gives the third bound. The two additional events have probability at most \(2e^{-\ell}\).

For the moment bound let \(M_+>0\). The inclusive quantile bound and Markov's inequality give \(r/2\le\sigma^{1+\varepsilon}/M_+^{1+\varepsilon}\), so \(rM_+\le2^{1/(1+\varepsilon)}\sigma r^{\gamma_\varepsilon}\). Also \(V_+^{(\bar\varepsilon)}\le M_+^{\bar\varepsilon-\varepsilon}\sigma^{1+\varepsilon}\), hence
\[
r^{\gamma_{\bar\varepsilon}}\bigl(V_+^{(\bar\varepsilon)}\bigr)^{1/(1+\bar\varepsilon)}\le2^{\frac{\bar\varepsilon-\varepsilon}{(1+\bar\varepsilon)(1+\varepsilon)}}\sigma r^{\gamma_\varepsilon}\le\sqrt2\,\sigma r^{\gamma_\varepsilon}.
\]
The third bound of \eqref{eq:moment-trimming} and the subadditivity of \(x\mapsto x^{1/(1+\bar\varepsilon)}\) give
\[
\widehat w^{(\bar\varepsilon)}\le\tfrac83\,r^{\gamma_{\bar\varepsilon}}\bigl(V_+^{(\bar\varepsilon)}\bigr)^{1/(1+\bar\varepsilon)}+\bigl(\tfrac{14}9+9\bigr)rM_+
\le\bigl[\tfrac83\sqrt2+2\bigl(\tfrac{14}9+9\bigr)\bigr]\sigma r^{\gamma_\varepsilon}<25\,\sigma r^{\gamma_\varepsilon}.
\]
If \(M_+=0\), the retained observations vanish and \(\widehat w^{(\bar\varepsilon)}=0\).
\end{proof}

\begin{lemma}[Centering]\label{lem:centering}
For every \(0<\varepsilon\le1\), every \(z\in\mathbb R\) and every set \(E\) of positive probability,
\[
\sigma_{E,\varepsilon}\le2\bigl(\mathbb E[|X-z|^{1+\varepsilon}\mathbf1_E]\bigr)^{1/(1+\varepsilon)}.
\]
At \(\varepsilon=1\) the factor \(2\) can be replaced by \(1\).
\end{lemma}
\begin{proof}
Jensen's inequality gives \((1-p)^{1/(1+\varepsilon)}|\mu_E-z|\le(\mathbb E[|X-z|^{1+\varepsilon}\mathbf1_E])^{1/(1+\varepsilon)}\), and Minkowski's inequality on \(E\) bounds \(\sigma_{E,\varepsilon}\) by the sum of these two quantities. The case \(\varepsilon=1\) is \eqref{eq:retained-variance}.
\end{proof}

\section{Proof of \Cref{thm:integrability}}\label{app:profiles}

Preassign to each arm an independent pilot sample and an independent fresh sequence \citep{lattimore2020}. Fix a suboptimal arm \(i\) and condition on all pilot medians. For \(n_0\le n\le T\) and each of the arms \(i\) and \(i^\star\), apply \Cref{lem:trimming} (if \(\bar\varepsilon=1\)) or \Cref{lem:trimming-moment} (if \(\bar\varepsilon<1\)) to the first \(n\) fresh observations shifted by the pilot median, with \(r=\ell/n\). Since \(e^{-\ell/6}\le T^{-4}\), all these events hold together except with probability \(18T^{-3}\), and \(14T^{-3}\) when \(\bar\varepsilon=1\). Each event concerns a fixed prefix of a pre-generated sequence, so it remains valid at the random sample sizes selected by the policy. The warm-up ensures \(n\ge n_0\ge4\ell\) and \(2r\le q\).

\textbf{Optimism at every pilot value.} Consider the optimal arm, with mean \(\mu^\star\), index \(U_\star\) and pilot median \(z\), and let \(E=\{|X-z|\le M_-\}\), whose omitted probability is \(p_-\le2r\le q\). If \(\bar\varepsilon=1\), \eqref{eq:retained-variance} gives \(v_E\le V_-\), so visibility and \eqref{eq:quantile-trimming} give
\[
\mu^\star-\mu_-=\frac{b_E}{1-p_-}\le\frac{a\sqrt{p_-V_-}}{1-p_-}\le3a\widehat w,\qquad
U_\star\ge\mu_--3\widehat w+3(1+a)\widehat w\ge\mu^\star.
\]
If \(\bar\varepsilon<1\), visibility, \Cref{lem:centering} and \eqref{eq:moment-trimming} give
\[
\begin{aligned}
b_E\le a\,p_-^{\gamma_{\bar\varepsilon}}\sigma_{E,\bar\varepsilon}
&\le2a(2r)^{\gamma_{\bar\varepsilon}}\bigl(2\widehat V^{(\bar\varepsilon)}+6r\widehat M^{1+\bar\varepsilon}\bigr)^{\frac1{1+\bar\varepsilon}}\\
&\le a\Bigl(4r^{\gamma_{\bar\varepsilon}}\bigl(\widehat V^{(\bar\varepsilon)}\bigr)^{\frac1{1+\bar\varepsilon}}+12r\widehat M\Bigr),
\end{aligned}
\]
so \(\mu^\star-\mu_-\le2b_E\le8a\widehat w^{(\bar\varepsilon)}\) and \(U_\star^{(\bar\varepsilon)}\ge\mu_--3\sqrt2\,\widehat w^{(\bar\varepsilon)}+8(1+a)\widehat w^{(\bar\varepsilon)}\ge\mu^\star\). No accuracy of the pilot was used.

\textbf{A competitor with a finite moment.} Fix \(\varepsilon\in(0,\bar\varepsilon]\) with \(\sigma=\sigma_{i,\varepsilon}<\infty\). If \(\sigma=0\), the arm is a point mass, its pilot is exact and its index equals \(\mu_i\). Otherwise Markov and Hoeffding give
\[
\mathbb P[|X-\mu_i|>4\sigma]\le1/4,\qquad
\mathbb P[|Z_i-\mu_i|>4\sigma]\le e^{-m/8}\le T^{-4}.
\]
On the complementary event, Minkowski gives \((\mathbb E[|X-Z_i|^{1+\varepsilon}])^{1/(1+\varepsilon)}\le5\sigma\). Write \(Y=X-Z_i\). Since
\[
\mu_--\mu_i=\frac{p_-\mathbb E[Y]-\mathbb E[Y\mathbf1_{\{|Y|>M_-\}}]}{1-p_-},
\]
Hölder gives \(|\mu_--\mu_i|\le10\sigma p_-^{\gamma_\varepsilon}/(1-p_-)\le40\sigma r^{\gamma_\varepsilon}\). With the moment bounds of the two trimming lemmas at scale \(5\sigma\),
\[
\begin{aligned}
U_i-\mu_i&\le5[8+75(2+a)]\sigma r^{\gamma_\varepsilon}\le800(1+a)\sigma r^{\gamma_\varepsilon},\\
U_i^{(\bar\varepsilon)}-\mu_i&\le5[8+25(3\sqrt2+8(1+a))]\sigma r^{\gamma_\varepsilon}\le1600(1+a)\sigma r^{\gamma_\varepsilon}.
\end{aligned}
\]
Only the competing arm whose pulls are being bounded needs the pilot event.

\textbf{Counting.} On these events a pull of arm \(i\) at fresh sample size \(n\) requires \(U_i\ge U_\star\ge\mu^\star\), hence \(\Delta_i\le800(1+a)\sigma(\ell/n)^{\gamma_\varepsilon}\). Arm \(i\) is pulled at most once at each fresh sample size, so
\begin{equation}\label{eq:central-count}
\mathbb E[N_i(T)]\le m+n_0+1+\frac{18}{T^2}+\frac1{T^3}
+\ell\left(\frac{800(1+a)\sigma}{\Delta_i}\right)^{1/\gamma_\varepsilon},
\end{equation}
with \(1600\) in place of \(800\) when \(\bar\varepsilon<1\). An unfinished warm-up satisfies the same bound directly. For the benchmark, \(m\le32\log T+3\) and \(n_0\le(50/q)\log T+1\), so \(m+n_0+1+18/T^2+T^{-3}\le(40+50/q)\log T\) for \(T\ge3\). Also \(\ell\le25\log T\) and \(25\le5^{1/\gamma_\varepsilon}\), which turns \(800\) and \(1600\) into \(4000\) and \(8000\).

\textbf{Profile count.} For \(\bar\varepsilon=1\), the last bound of \eqref{eq:quantile-trimming} gives \(U_i\le\mu_-+3(2+a)\widehat w\le\mu_i+H_{\nu_i-Z_i,a}(\ell/n)\) on the same events, with no moment assumption. Conditionally on the pilots,
\begin{equation}\label{eq:profile-count}
\mathbb E[N_i(T)]\le m+n_0+
\sum_{n=n_0}^{T-1}\mathbf1\{\Delta_i\le H_{\nu_i-Z_i,a}(\ell/n)\}+14/T^2.
\end{equation}
No monotonicity of \(H\) is required.

\subsection{Finite means and endpoint envelopes}\label{app:finite-means}

For integrable \(X\), write \(M_+=M_+(r)\). Since \(\mathbb P[|X|\ge M_+]\ge r/2\),
\[
rM_+\le2M_+\mathbb P[|X|\ge M_+]\longrightarrow0,
\qquad rV(M_+)\le rM_+\mathbb E[|X|]\longrightarrow0.
\]
When \(M_+\) remains bounded the limits are immediate. Otherwise use \(t\mathbb P[|X|\ge t]\le\mathbb E[|X|\mathbf1_{\{|X|\ge t\}}]\to0\). Absolute continuity of the integral also gives \(B_\nu(r)\to0\), since its omitted set has probability at most \(2r\). Thus \(H_{\nu,c}(r)\to0\). These arguments also cover bounded support and atoms, where a quantile need not diverge. The first-moment assumption enters through absolute continuity of the integral.

For centered pilots, define \(M_i(r)=\inf\{M:\mathbb P[|X-\mu_i|>M]\le r/2\}\). Set \(\tau_i=\mathbb E[|X-\mu_i|]\), \(\bar M_i(r)=M_i(r)+4\tau_i\), and
\[
\omega_i(p)=\sup_{\mathbb P[F]\le p}\mathbb E[|X-\mu_i|\mathbf1_F].
\]
Except with probability \(T^{-4}\), \(|Z_i-\mu_i|\le4\tau_i\). Uniformly over these centers, the shifted upper quantile is at most \(\bar M_i(r)\) and its first absolute moment is at most \(5\tau_i\). Therefore its profile is bounded by
\[
\overline H_i(r)=\frac{\omega_i(2r)}{1-2r}
+3(2+a)\left\{3\sqrt{5\tau_i r\bar M_i(r)}+11r\bar M_i(r)\right\}\longrightarrow0.
\]
The shifted lower-quantile omission has probability at most \(2r\). Since the central residual has mean zero, its conditional mean displacement is its omitted central integral divided by the retained probability. This proves the first term in \(\overline H_i\). The width bound uses \(V_z(M)\le M\mathbb E[|X-z|]\). Finally \(rM_i(r)\to0\) by the preceding quantile argument, while \(\omega_i(p)\to0\) by absolute continuity. All convergence statements are uniform over the stated interval of pilot centers.

A point-mass competitor has zero profile. Conditional application of \eqref{eq:profile-count} proves
\[
\mathbb E[N_i(T)]\le m+n_0+
\sum_{n=n_0}^{T-1}\mathbf1\{\Delta_i\le\overline H_i(\ell/n)\}
+14/T^2+T^{-3}.
\]
Choose \(r_i>0\) with \(\overline H_i(r)<\Delta_i\) for \(r\le r_i\). The sum has at most \(\ell/r_i+1\) terms, proving logarithmic regret with only finite first moments. For \(\bar\varepsilon<1\) the same argument applies to the index \(U_i^{(\bar\varepsilon)}\): by \eqref{eq:moment-trimming} and \(V_+^{(\bar\varepsilon)}\le5\tau_i\bar M_i(r)^{\bar\varepsilon}\), its excess over \(\mu_i\) is at most
\[
\frac{\omega_i(2r)}{1-2r}+\bigl(3\sqrt2+8(1+a)\bigr)\Bigl[\bigl(\tfrac{40}3\tau_i\bigr)^{\frac1{1+\bar\varepsilon}}\bigl(r\bar M_i(r)\bigr)^{\gamma_{\bar\varepsilon}}+11r\bar M_i(r)\Bigr]\longrightarrow0.
\]

For an envelope \(\mathbb P[|X|>t]\le At^{-\alpha}\), put \(\sigma_A=A^{1/\alpha}\) and recall \(\gamma_{\alpha-1}=(\alpha-1)/\alpha\). Tail integration gives
\[
M_+\le(2A/r)^{1/\alpha},\quad M_-\le(A/(2r))^{1/\alpha},\quad
V(M)\le\frac{2A}{2-\alpha}M^{2-\alpha},
\]
\[
\mathbb E[|X|\mathbf1_F]\le\int_0^\infty\min\{\mathbb P[F],At^{-\alpha}\}\,dt
=\frac\alpha{\alpha-1}\sigma_A\mathbb P[F]^{\gamma_{\alpha-1}}.
\]
Define the explicit constants
\[
G_\alpha=2^{(\alpha-1)/\alpha}\left(\frac\alpha{\alpha-1}+1\right),\quad
J_\alpha=2^{1/\alpha}\left(\frac3{\sqrt{2-\alpha}}+11\right),\quad
D_\alpha(c)=G_\alpha+3(2+c)J_\alpha.
\]
The identity \(\mu_- -\mu=p_-\mu_- -\mathbb E[X\mathbf1_{\{|X|>M_-\}}]\) now yields
\[
B_\nu(r)\le G_\alpha\sigma_Ar^{\gamma_{\alpha-1}},\quad W_\nu(r)\le J_\alpha\sigma_Ar^{\gamma_{\alpha-1}},\quad
H_{\nu,c}(r)\le D_\alpha(c)\sigma_Ar^{\gamma_{\alpha-1}}.
\]
Counting in \eqref{eq:profile-count} proves \Cref{cor:endpoint}. For a central envelope, the pilot event \(|Z_i-\mu_i|\le(4A)^{1/\alpha}\) has failure probability at most \(T^{-4}\). On it, \(\mathbb P[|X-Z_i|>t]\le16At^{-\alpha}\), using \(|X-\mu_i|>t/2\) when \(t>2(4A)^{1/\alpha}\) and the bound one otherwise. Apply the raw result conditionally with \(16A\).

\section{Proof of the nesting in \Cref{thm:scale}}\label{app:scale}

Fix \(E\) with \(0<p\le q/2\) and \(\sigma_{E,\varepsilon}<\infty\). If \(b_E\le0\) there is nothing to prove, and if \(\sigma_{E,\varepsilon}=0\) then \(v_E=0\) and variance-core visibility gives \(b_E\le0\). Let \(b_E>0\) and \(\sigma_{E,\varepsilon}>0\), and translate so that \(\mu_E=0\), which gives \(b_E=(1-p)\mu\) with \(\mu>0\). Put \(t=\sigma_{E,\varepsilon}p^{-1/(1+\varepsilon)}\) and \(F=E\cap\{|X|\le t\}\). Markov's inequality on \(E\) gives \(\mathbb P[E\setminus F]\le\sigma_{E,\varepsilon}^{1+\varepsilon}/t^{1+\varepsilon}=p\), so the omitted probability \(p_F\) of \(F\) satisfies \(p\le p_F\le2p\le q\). Also
\[
v_F\le\mathbb E[X^2\mathbf1_F]\le t^{1-\varepsilon}\sigma_{E,\varepsilon}^{1+\varepsilon},\qquad
\bigl|\mathbb E[X\mathbf1_{E\setminus F}]\bigr|\le t^{-\varepsilon}\sigma_{E,\varepsilon}^{1+\varepsilon}=\sigma_{E,\varepsilon}p^{\gamma_\varepsilon}.
\]
Because \(\mathbb E[X\mathbf1_E]=0\), we have \(b_F=(1-p_F)\mu+\mathbb E[X\mathbf1_{E\setminus F}]\). Variance-core visibility applied to \(F\) gives
\[
b_F\le a\sqrt{p_Fv_F}\le a\sqrt{2p\,t^{1-\varepsilon}\sigma_{E,\varepsilon}^{1+\varepsilon}}=\sqrt2\,a\,\sigma_{E,\varepsilon}p^{\gamma_\varepsilon},
\]
hence \((1-p_F)\mu\le(\sqrt2a+1)\sigma_{E,\varepsilon}p^{\gamma_\varepsilon}\). Finally \(b_E=(1-p)\mu\) and \((1-p)/(1-p_F)\le(1-p)/(1-2p)\le3/2\) for \(p\le1/4\).

\section{Contamination, instance bounds and consequences}\label{app:contamination}

\begin{proof}[Proof of \Cref{lem:contamination}]
The mean is immediate, and \(\nu\ge(1-\varpi)\pi\) gives the divergence bound. Fix a set \(E\) whose omitted probability \(p\) under \(\nu\) satisfies \(0<p\le q\), with finite retained variance. Write \(p'\), \(b'\) and \(v'\) for the omitted probability, the omitted centered mean and the retained variance of \(E\) under \(\pi\), and \(b_E,v_E\) for those under \(\nu\). Minimizing over the center gives \(v_E\ge(1-\varpi)v'\). Let \(x=\mu_\pi+\iota/\varpi\) be the new atom.

If \(x\in E\), then \(p=(1-\varpi)p'\), so \(0<p'\le2q\), and
\[
b_E=(1-\varpi)(b'-p'\iota)\le(1-\varpi)b'\le a'(1-\varpi)\sqrt{p'v'}\le a\sqrt{pv_E}.
\]
If \(x\notin E\), then \(p=\varpi+(1-\varpi)p'\) with \(p'\le q\), the retained part of \(\nu\) is \((1-\varpi)\pi\) restricted to \(E\), so \(v_E=(1-\varpi)v'\), and
\[
b_E=\varpi(x-\mu_\pi-\iota)+(1-\varpi)(b'-p'\iota)=(1-\varpi)\bigl[b'+(1-p')\iota\bigr].
\]
Visibility of \(\pi\) gives \(b'\le a'\sqrt{p'v'}\), and the choice \(F=E\) in the capacity gives \(\iota\le(a-a')\sqrt{(\varpi+p')v'}\). Both terms are controlled by \(\sqrt{(1-\varpi)p}\), because
\[
(1-\varpi)^2p'\le(1-\varpi)p,\qquad(1-\varpi)^2(1-p')^2(\varpi+p')\le(1-\varpi)(\varpi+p'-\varpi p')=(1-\varpi)p.
\]
Therefore \(b_E\le a\sqrt{(1-\varpi)p\,v'}=a\sqrt{pv_E}\). If \(v'=0\), the same computation gives \(b'\le0\) and \(\iota=0\), so \(b_E\le0\).
\end{proof}

\begin{proof}[Proof of \Cref{cor:instance}]
\textup{(i)} The definition gives \(\Gamma_\pi(\varpi)\ge\sqrt{\varpi\,\underline v_q}\). Take \(\varpi=4\Delta^2/[(a-a')^2\underline v_q]\le q/2\). Then \((a-a')\Gamma_\pi(\varpi)\ge2\Delta\) and \(\log\frac1{1-\varpi}\le2\varpi\), and \Cref{thm:instance} gives the first bound. Let \(\pi\) have variance \(\sigma^2=\sigma_1(\pi)^2\) and let \(\bar\omega(p)=\sup\{\mathbb E_\pi[(X-\mu_\pi)^2\mathbf1_F]:\pi(F)\le p\}\), which tends to zero with \(p\). For a set \(F\) with \(p'=\pi(F^c)\le p\), Cauchy--Schwarz gives \(b_F^2\le p'\bar\omega(p)\), so
\[
v_F=\mathbb E_\pi[(X-\mu_\pi)^2\mathbf1_F]-\frac{b_F^2}{1-p'}\ge\sigma^2-\frac{\bar\omega(p)}{1-p}.
\]
For \(p'>p\), \((\varpi+p')v_F\ge p\,\underline v_q\). Hence \(\Gamma_\pi(\varpi)^2\ge\min\{\varpi(\sigma^2-\bar\omega(p)/(1-p)),\,p\,\underline v_q\}\), and \(F=\mathbb R\) gives \(\Gamma_\pi(\varpi)^2\le\varpi\sigma^2\). Letting \(\varpi\downarrow0\) and then \(p\downarrow0\) gives \(\Gamma_\pi(\varpi)\sim\sigma\sqrt\varpi\), and the weight \(\varpi=[(1+\delta)\Delta/((a-a')\sigma)]^2\) in \Cref{thm:instance}, followed by \(\delta\downarrow0\), gives the second bound.

\textup{(ii)} For \(p'>0\) we have \((\varpi+p')p'^{-(2-\alpha)/\alpha}\ge\varpi^{2(\alpha-1)/\alpha}\): if \(p'\le\varpi\) the second factor is at least \(\varpi^{-(2-\alpha)/\alpha}\), and otherwise the product is at least \(p'^{2(\alpha-1)/\alpha}\). The term \(F=\mathbb R\) is infinite. So \(\Gamma_\pi(\varpi)^2\ge\underline v_\alpha\varpi^{2(\alpha-1)/\alpha}\). The weight \(\varpi=[2\Delta/((a-a')\sqrt{\underline v_\alpha})]^{\alpha/(\alpha-1)}\), which is at most \(q/2\) for small \(\Delta\), gives \((a-a')\Gamma_\pi(\varpi)\ge2\Delta\), and \(\log\frac1{1-\varpi}\le2\varpi\) gives the claim.
\end{proof}

\textbf{A Pareto mixture.} Condition (i) holds for every non-degenerate law once \(q\) is small, and can fail at a fixed \(q\): for \(\pi=(1-p)\delta_0+p\delta_{-1}\) with \(p\le q\), retaining \(\{0\}\) gives \(\underline v_q=0\). For condition (ii), let \(1<\alpha<2\), \(\varpi_\alpha=(\alpha-1)/(8\alpha)\) and \(q\le\varpi_\alpha/4\), and let
\[
\pi=(1-\varpi_\alpha)\delta_0+\varpi_\alpha\,\mathcal L(-X_\alpha),\qquad X_\alpha\sim\mathrm{Pareto}(\alpha,x_0).
\]
The mean is \(-\varpi_\alpha\alpha x_0/(\alpha-1)=-x_0/8\) and every negative reward is at most \(-x_0\). The atom at \(0\) has mass above \(2q\), so every admissible omission removes only rewards below the mean, and \(\pi\in\mathrm{VCV}(0,2q)\). Let \(F\) have omitted probability \(p'\in(0,q]\). It retains \(0\), and Cauchy--Schwarz gives \((\mathbb E_\pi[X\mathbf1_F])^2\le(\varpi_\alpha-p')\mathbb E_\pi[X^2\mathbf1_F]\). Since \((\varpi_\alpha-p')/(1-p')\le\varpi_\alpha\), this gives \(v_F\ge(1-\varpi_\alpha)\mathbb E_\pi[X^2\mathbf1_F]\). With \(A=\varpi_\alpha x_0^\alpha\) we have \(\pi(X<-y)=Ay^{-\alpha}\) for \(y\ge x_0\). Put \(M=(A/(4p'))^{1/\alpha}\ge x_0\). The band \(M<|X|\le2^{1/\alpha}M\) has probability \(2p'\), at least \(p'\) of it remains in \(F\), and so \(\mathbb E_\pi[X^2\mathbf1_F]\ge p'M^2\). Condition (ii) therefore holds with \(\underline v_\alpha=(1-\varpi_\alpha)(A/4)^{2/\alpha}\).

\begin{proof}[Proof of \Cref{prop:consequences}]
\textup{(i)} Fix \(\Delta>0\). Arms with gap at most \(\Delta\) contribute at most \(\Delta T\), and the others at most \(C_0\log T\sum_i\Delta_i+K(C_1\sigma)^{1/\gamma_\varepsilon}\Delta^{1-1/\gamma_\varepsilon}\log T\). The choice \(\Delta=C_1\sigma(K\log T/T)^{\gamma_\varepsilon}\) makes both remaining terms equal to \(C_1\sigma T^{1-\gamma_\varepsilon}(K\log T)^{\gamma_\varepsilon}\).

\textup{(ii)} Let epoch \(j\) have length \(T_j=3^{2^j}\). Regret over a prefix of an epoch is at most regret over the epoch. If \(T\) lies in epoch \(j\ge1\), then \(T>T_{j-1}\), so \(\log T_j=2\log T_{j-1}<2\log T\), and \(\sum_{k\le j}\log T_k<2\log T_j\). With known inputs, the restarted algorithm therefore has regret at most four times the benchmark of \Cref{thm:integrability} at every \(T\ge3\). With the inputs of the statement, the classes \(\mathrm{VCV}(\Phi(t),1/(2\Phi(t)))\) increase with \(t\), so the inputs are valid from some epoch on, and the earlier epochs contribute a bounded regret. Every epoch started by time \(T\) has \(1+\Phi(t)\le2\Phi(T)\), and \Cref{thm:integrability} gives the bound.

\textup{(iii)} Suppose \(\mathbb E_0[N_2(T)]\le n_T:=C\log T\) for large \(T\) on the instance \((\delta_1,\delta_0)\). Choose \(p\in(0,1)\) with \(\eta=2C\log\frac1{1-p}<1\) and let \(\nu=(1-p)\delta_0+p\,\delta_{2/p}\), which has mean \(2\) and lies in \(\mathrm{VCV}(0,q)\) for every \(q<p\). The likelihood argument of \Cref{thm:frontier}\textup{(ii)}, applied to \((\delta_1,\delta_0)\) and \((\delta_1,\nu)\) with \(F_T=\{N_2(T)\le2n_T\}\), gives \(R_T(\delta_1,\nu)\ge\frac14T^{1-\eta}\) for large \(T\).
\end{proof}

\section{The raw-moment formulation}\label{app:raw}

Let \(\sigma^\circ_\varepsilon(\nu)=(\mathbb E_\nu[|X|^{1+\varepsilon}])^{1/(1+\varepsilon)}\) be the raw scale. The raw budget and the raw benchmark \(\mathcal B_T^\circ\) replace \(\sigma_\varepsilon\) by \(\sigma^\circ_\varepsilon\) in \Cref{def:budget} and in \(\mathcal B_T\). Raw VT-UCB omits the pilot stage and fixes every \(Z_i=0\), with threshold-visibility inputs \((c,q)\) in place of \((a,q)\). This appendix treats the variance budget, that is, the exponent \(\varepsilon=1\).

\begin{proposition}[Raw frontier]\label{prop:raw}
\textup{(i)} If the optimal law satisfies \(X\in\mathrm{VT}(c,q)\) and every arm is integrable, raw VT-UCB satisfies \(R_T\le\mathcal B_T^\circ(4+50/q,\,800(1+c);1)\).

\textup{(ii)} Let a strategy have raw variance budget \((C_0,C_1)\). Let \(\nu\) be an integrable law and \(E\) a set with \(0<p\le\min\{1/2,1/(8C_0)\}\) and \(b_E>4C_1\sqrt{p\,\mathbb E[X^2\mathbf1_E]}\). Then the conclusion of \Cref{thm:frontier}\textup{(ii)} holds with \(\mathbb E[X^2\mathbf1_E]\) in place of \(v_E\). By \Cref{app:converse}, a law with no such set satisfies \(X\in\mathrm{VT}(4C_1,\min\{1/2,1/(8C_0)\})\).

\textup{(iii)} If a strategy has raw variance budget \((C_0,C_1)\) and is uniformly efficient on every two-armed instance whose optimal law satisfies \(X\in\mathrm{VT}(c,q)\), then \(C_0\ge1/\log\frac1{1-q}\).
\end{proposition}
\begin{proof}
\textup{(i)} Repeat the proof of \Cref{thm:integrability} with every center equal to zero. For the optimal arm, \eqref{eq:quantile-trimming} gives \(\mu^\star-\mu_-\le cs(M_-)/(1-p_-)\le3c\widehat w\), so \(U_\star\ge\mu^\star\). For a competitor with raw scale \(\sigma=\sigma^\circ_\varepsilon(\nu_i)\) there is no pilot event, and \(U_i-\mu_i\le[8+75(2+c)]\sigma r^{\gamma_\varepsilon}\le160(1+c)\sigma r^{\gamma_\varepsilon}\). Counting gives \(\mathbb E[N_i(T)]\le n_0+1+14/T^2+\ell(160(1+c)\sigma/\Delta_i)^{1/\gamma_\varepsilon}\), and \(n_0+1+14/T^2\le(4+50/q)\log T\). \textup{(ii)} The conditional law has raw second moment \(\mathbb E[X^2\mathbf1_E]/(1-p)\), and the algebra of \Cref{thm:frontier}\textup{(ii)} is unchanged. \textup{(iii)} Repeat the allowance part of the proof of \Cref{thm:frontier}\textup{(iii)}. A point mass, and the law \((1-p)\delta_0+p\delta_1\) with \(p>q\), have no threshold \(M\) with \(0<p(M)\le q\), so both satisfy \(X\in\mathrm{VT}(0,q)\), and the competitor \(\delta_0\) has raw scale zero.
\end{proof}

\textbf{Separation from the centered frontier.} Let \(\Pi_p=(1-p)\delta_{p^{-1/2}}+p\delta_{p^{-1/2}+1/p}\) for \(0<p\le1/2\). At thresholds between its atoms, \(b=1-p\) and \(V=(1-p)/p\), so \(\Pi_p\in\mathrm{VT}(1,1/2)\) uniformly. Yet retaining the lower atom gives \(v_E=0\) and \(b_E=1-p>0\), so no class \(\mathrm{VCV}(a,q)\) contains the family. The raw scale pays for a large artificial location.

\section{The exponential coefficient}\label{app:gap}

Recall \(x_\Delta=2A-\Delta\), where \(A=e/10\). Tail integration gives \(\mathbb E[\xi]=2A\) and \(\mathbb E[\xi^{1+\varepsilon}]=\infty\) for every \(\varepsilon>0\). Every omitted reward set of probability at most \(1/4\) excludes the atom at \(x_\Delta\), whose mass is \(9/10\). Its remaining rewards lie below \(x_\Delta-e<-\Delta\). Thus \(P_\Delta\) is VCV\((0,1/4)\).

For \(X\sim P_\Delta\), define
\[
f(\lambda)=\mathbb E[\log(1-\lambda X)],\qquad0\le\lambda<1/x_\Delta.
\]
Strict concavity, \(f'(0)=\Delta\) and \(f'(\lambda)\to-\infty\) at \(1/x_\Delta\) give a unique maximizer \(\lambda_\Delta\). For \(\lambda>\lambda_\Delta\) sufficiently close, define
\[
\Omega_\lambda=\mathbb E[(1-\lambda X)^{-1}],\qquad
\frac{dQ_\lambda}{dP_\Delta}(x)=\frac1{\Omega_\lambda(1-\lambda x)}.
\]
Since \(\Omega_\lambda=1+\lambda\mathbb E[X/(1-\lambda X)]\), \(Q_\lambda\) has positive mean, while \(\Omega_{\lambda_\Delta}=1\). The logarithm defining \(f\) is integrable, since its negative-tail part is bounded by \(\lambda|X|\) and its positive support is bounded. Differentiation is justified on every compact subinterval of \((0,1/x_\Delta)\). Strict concavity follows from the nondegenerate law, and the positive atom forces the derivative to diverge at the right endpoint. The variance of \(Q_\lambda\) is finite because \(x^2/(1-\lambda x)\le|x|/\lambda\) on the negative tail. Moreover
\[
(1-\lambda x_\Delta)\Omega_\lambda\le1,
\]
so its positive atom has probability at least \(9/10\). The same omitted-set argument gives VCV\((0,1/4)\). Finally,
\[
\operatorname{KL}(P_\Delta\Vert Q_\lambda)=f(\lambda)+\log \Omega_\lambda
\longrightarrow d_\Delta:=f(\lambda_\Delta).
\]
The change-of-measure argument in \Cref{app:converse}, applied to \((\delta_0,P_\Delta)\) and \((\delta_0,Q_\lambda)\), proves the lower coefficient \(\Delta/d_\Delta\).

To evaluate it, put \(\Lambda=\log(1/\lambda)\) and
\[
g_\Delta(\lambda)=\lambda\mathbb E\left[\frac{(\xi-x_\Delta)^2}{1+\lambda(\xi-x_\Delta)}\right].
\]
Uniformly for \(0<\Delta<A/2\), replacing \(\xi-x_\Delta\) by \(\xi\) changes this by \(O(\lambda)\). The derivative of \(\phi_\lambda(w)=\lambda w^2/(1+\lambda w)\) is \(1-(1+\lambda w)^{-2}\), bounded on the intervening segment by a constant multiple of \(\lambda(\xi+x_\Delta)\). Tail integration and \(y=e^t\) give
\[
g_\Delta(\lambda)=A\int_1^\infty\frac{k(t-\Lambda)}{t^2}\,dt+O(\lambda),\qquad
k(\varsigma)=\frac{e^\varsigma(2+e^\varsigma)}{(1+e^\varsigma)^2}.
\]
Writing \(\chi(\varsigma)=e^\varsigma/(1+e^\varsigma)\) gives \(k=\chi+\chi'\). Thus \(\psi=k-\mathbf1_{\{\varsigma>0\}}\) has exponential tails and integral one, since \(\chi(-\varsigma)=1-\chi(\varsigma)\). Expand \((\Lambda+\varsigma)^{-2}\) on \(|\varsigma|\le\Lambda/2\). The correction is \(\Lambda^{-2}+O(\Lambda^{-3})\), while both complementary integrals are exponentially small. Hence
\begin{equation}\label{eq:tilt-expansion}
g_\Delta(\lambda)=\frac A\Lambda+\frac A{\Lambda^2}+O(\Lambda^{-3}).
\end{equation}
The maximizing parameter tends to zero as \(\Delta\) tends to zero. Otherwise \(g_\Delta(\lambda)\) would stay bounded away from zero along a subsequence, contradicting its equality to \(\Delta\). Uniformity holds because \(x_\Delta\) stays in the compact interval \([3A/2,2A]\). The equation \(f'(\lambda_\Delta)=\Delta-g_\Delta(\lambda_\Delta)=0\) yields
\[
\Lambda_\Delta=A/\Delta+1+O(\Delta),\qquad
\lambda_\Delta=e^{-1}e^{-A/\Delta}[1+O(\Delta)].
\]
Using \(\tilde\lambda=\lambda e^{-\tau}\) in the integral of \eqref{eq:tilt-expansion} gives
\[
\int_0^\lambda g_\Delta(\tilde\lambda)\,d\tilde\lambda
=\lambda A[\Lambda^{-1}+O(\Lambda^{-3})],
\]
because
\[
\int_0^\infty\frac{e^{-\tau}}{\Lambda+\tau}\,d\tau=\Lambda^{-1}-\Lambda^{-2}+O(\Lambda^{-3}),\qquad
\int_0^\infty\frac{e^{-\tau}}{(\Lambda+\tau)^2}\,d\tau=\Lambda^{-2}+O(\Lambda^{-3}).
\]
These expansions follow by Taylor expansion under the exponentially integrable weight, with the remainder bounded by its second moment. Their second-order terms cancel. Consequently
\[
d_\Delta=\lambda_\Delta\Delta-\int_0^{\lambda_\Delta}g_\Delta(\tilde\lambda)\,d\tilde\lambda
=\frac{A\lambda_\Delta}{\Lambda_\Delta^2}[1+O(\Lambda_\Delta^{-1})]
\sim\frac{\Delta^2}{Ae}e^{-A/\Delta}.
\]

For the upper bound, the pilot median equals \(x_\Delta\) unless at least half its observations come from the component of mass \(1/10\). Hoeffding bounds this failure by \(T^{-4}\). Conditional on success, fresh shifted rewards have law \(-\xi\). Write \(y_-=e^{L_-}\) and \(y_+=e^{L_+}\) for the quantiles of \(\xi\) at tail probabilities \(2r\) and \(r/2\). Then
\[
e^{L_-}L_-^2=A/(2r),\qquad e^{L_+}L_+^2=2A/r,\qquad L_+-L_-\to\log4.
\]
Tail integration gives
\[
\mathbb E[\xi\mathbf1_{\{\xi>y\}}]=A/\log y+A/(\log y)^2,
\]
\[
V(y)=2\int_0^y v\mathbb P[\xi>v]\,dv-y^2\mathbb P[\xi>y]
=\frac{Ay}{(\log y)^2}[1+O((\log y)^{-1})].
\]
The last expansion follows by integration by parts in \(\int_e^y(\log v)^{-2}dv\). The bias is exactly
\[
B(r)=\frac{\mathbb E[\xi\mathbf1_{\{\xi>y_-\}}]-2A\mathbb P[\xi>y_-]}{1-\mathbb P[\xi>y_-]}.
\]
Since the omitted probability is exponentially small in \(L_-\), it follows that
\[
B(r)=A/L_-+A/L_-^2+o(L_-^{-2}),\quad
y_+r\sim2A/L_+^2,\quad \sqrt{V(y_+)r}\sim\sqrt2A/L_+^2,
\]
\[
H_{-\xi,0}(r)=A/L_-+\kappa A/L_-^2+o(L_-^{-2}),\qquad\kappa=133+18\sqrt2.
\]
Set \(L_0=A/\Delta+\kappa+1\) and \(r_\Delta=A/(2e^{L_0}L_0^2)\). For sufficiently small \(\Delta\), the expansion and the decreasing majorant \(A/L+(\kappa+1/2)A/L^2\) imply \(H_{-\xi,0}(r)<\Delta\) for all \(r\le r_\Delta\). Therefore
\[
\frac1{r_\Delta}\sim2Ae^{\kappa+1}\Delta^{-2}e^{A/\Delta}.
\]
The profile bound, \(\ell/\log T\to24\), and the additional pilot cost \(m\Delta+\Delta T^{-3}\) prove the upper coefficient \(O(\Delta^{-1}e^{A/\Delta})\).

\subsection{Other stretched-exponential powers}\label{app:stretched}

\begin{corollary}[A continuum of gap laws]\label{cor:continuum}
For any fixed \(\beta>0\), choose \(A>0\) with \(A/e\le1/10\) and \(A(1+1/\beta)<e/2\). Replacing the survival function by \(A/[y(\log y)^{1+\beta}]\) and setting \(P_\Delta=\mathcal L(\mathbb E[\xi]-\Delta-\xi)\) gives, in the comparison of \Cref{thm:gap}, upper and lower logarithmic coefficients of the same order
\[
\Delta^{-1/\beta}\exp\!\left[\left(\frac A{\beta\Delta}\right)^{1/\beta}\right].
\]
Comparison constants depend on \(A,\beta\), and the visibility inputs stay \((0,1/4)\).
\end{corollary}

Fix \(\beta>0\) and choose \(A/e\le1/10\) and \(\mu_\beta=A(1+1/\beta)<e/2\). Let the survival function above \(e\) be \(A/[y(\log y)^{1+\beta}]\), with probability \(A/e\) on the positive component and the remaining mass at zero. For \(0<\Delta<\mu_\beta/2\), use \(X=\mu_\beta-\Delta-\xi\). Tail integration gives \(\mathbb E[\xi]=\mu_\beta\), while every higher moment diverges. The positive atom and the location of the remaining support prove VCV\((0,1/4)\) for this law and its rationally tilted alternatives exactly as above.

The calculation above replaces \(t^{-2}\) by \(t^{-1-\beta}\). The indicator part of \(k\) integrates to \(\Lambda^{-\beta}/\beta\), and \(\psi\), which has integral one, contributes \(\Lambda^{-\beta-1}\). Exponential tails bound the Taylor remainder uniformly. Consequently
\[
g_\Delta(\lambda)=\frac A\beta\Lambda^{-\beta}+A\Lambda^{-\beta-1}+O(\Lambda^{-\beta-2}).
\]
Writing \(L_\beta=[A/(\beta\Delta)]^{1/\beta}\), inversion of \(\Delta=g_\Delta(\lambda_\Delta)\) and the same cancellation of second-order terms give
\[
\log(1/\lambda_\Delta)=L_\beta+1+O(L_\beta^{-1}),\qquad
 d_\Delta\sim\frac A e L_\beta^{-\beta-1}e^{-L_\beta}.
\]
For the learner, integration by parts gives truncated second moment \(Ay/(\log y)^{1+\beta}[1+O(1/\log y)]\). The omitted first moment is exactly \(A L^{-\beta}/\beta+A L^{-\beta-1}\). Thus
\[
H(r)=\frac A\beta L_-^{-\beta}+\kappa A L_-^{-\beta-1}+o(L_-^{-\beta-1}),\qquad
 e^{L_-}L_-^{\beta+1}=A/(2r).
\]
Choosing \(L_-=L_\beta+\kappa+1\) and repeating the profile count proves that both coefficients have order
\[
\Theta_{A,\beta}\left(\Delta^{-1/\beta}
\exp\left[\left(\frac A{\beta\Delta}\right)^{1/\beta}\right]\right).
\]
The lower prefactor is asymptotic to \(eA^{1/\beta}\beta^{-1-1/\beta}\), and any upper prefactor larger than \(48e^{\kappa+1}A^{1/\beta}\beta^{-1-1/\beta}\) is valid for sufficiently small gaps. Pilot failures and initialization have the same negligible contributions as before.

\section{Comparison table and structural examples}\label{app:examples}

\begin{table}[!ht]
\centering
\small
\begin{tabular}{@{}>{\raggedright\arraybackslash}p{2.5cm}>{\raggedright\arraybackslash}p{3.2cm}>{\raggedright\arraybackslash}p{1.6cm}>{\raggedright\arraybackslash}p{2.7cm}>{\raggedright\arraybackslash}p{3.4cm}@{}}
\toprule
Work & Optimal-arm condition & Inputs & Competitor moment & Regret per arm\\
\midrule
\citet{bubeck2013} & moment of order \(1+\varepsilon\) & \(\varepsilon,u\) & order \(1+\varepsilon\) & \((u/\Delta)^{1/\varepsilon}\log T\)\\
\citet{genalti2024} & truncated non-positivity & none & raw, order \(1+\varepsilon\) & \([(u/\Delta)^{1/\varepsilon}+\Delta/\mathbb P[X\ne0]]\log T\)\\
\citet{tamas2025} & symmetry about the mean, on every arm & none & some higher moment & gap-dependent \(\log^2T\)\\
\citet{ashutosh2021} & some higher moment & scaling function & some higher moment & superlogarithmic\\
\citet{honda2015} & moment generating function near zero & upper endpoint & bounded above & \(\Delta\log T/D_{\inf}\)\\
\midrule
\Cref{thm:integrability} & \(\mathcal V_{\bar\varepsilon}(a,q)\) & \(a,q,\bar\varepsilon,T\) & central, order \(1+\varepsilon\le1+\bar\varepsilon\) & \([(u/\Delta)^{1/\varepsilon}+\Delta]\log T\)\\
\Cref{thm:integrability} & \(\mathcal V_{\bar\varepsilon}(a,q)\) & \(a,q,\bar\varepsilon,T\) & first moment & instance-dependent \(\log T\)\\
\bottomrule
\end{tabular}
\caption{Expected-regret guarantees for heavy-tailed stochastic bandits, organized by the condition on the optimal arm, the learner's inputs, and the requirement on competitors, with \(u\) a bound on the moment of order \(1+\varepsilon\). Here \(D_{\inf}\) is the smallest KL divergence from the competitor's law to a law whose mean exceeds the optimum. The restart scheme of \Cref{prop:consequences} removes the input \(T\) at the cost of a factor four.}
\label{tab:comparison}
\end{table}

\Cref{tab:comparison} compares the guarantees discussed in \Cref{sec:related}. The rest of this appendix collects exact examples.

\begin{figure}[!ht]
\centering
\includegraphics[width=\textwidth]{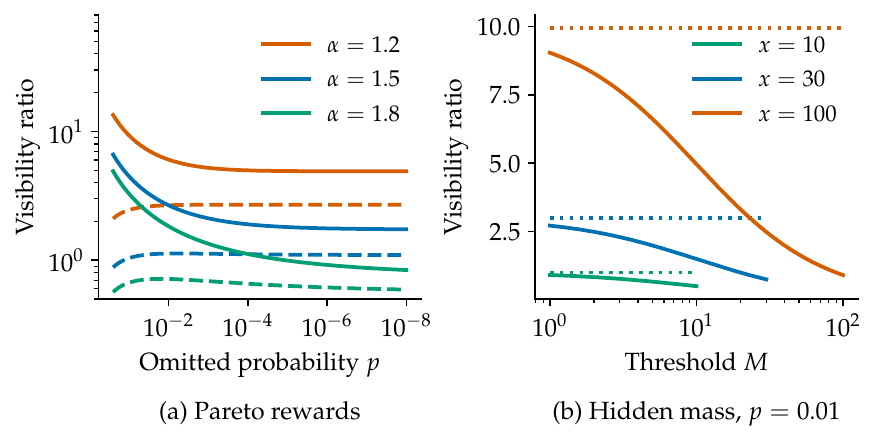}
\caption{Exact visibility ratios. (a) Pareto rewards against the omitted probability \(p\): variance-core ratio along \(E=\{X\le t\}\) (solid) and raw ratio \(\rho\) (dashed). (b) Hidden mass with \(p=0.01\): raw ratio \(\rho(M)\) (solid) and the variance-core value \(x\sqrt{p(1-p)}\) (dotted).}
\label{fig:visibility}
\end{figure}

\textbf{Hidden mass: see it or cover it.} Let \(X\) follow \((1-p)\mathrm U\{-1,1\}+p\delta_x\), with \(\mathrm U\{-1,1\}\) uniform on \(\{-1,1\}\), \(x>1\) and \(p\le1/3\). If \(q<p\), no set has omitted mass in \((0,q]\), so the law lies in \(\mathrm{VCV}(0,q)\), and the warm-up of size \(2\ell/q\) observes the atom. If \(q\ge p\), omitting the atom gives \(v_E=1-p\) and \(b_E=px(1-p)\), which forces \(a\ge x\sqrt{p(1-p)}\), so the width must cover a reward the sample can miss (\Cref{fig:visibility}b).

\textbf{Bounded laws and truncated non-positivity.} For \(p\le q\), the law Bernoulli\((p)\) fails \(\mathrm{VCV}(a,q)\) for every \(a\), because retaining \(\{0\}\) gives \(v_E=0<b_E\). By \Cref{thm:frontier}(ii), every strategy with a variance budget whose allowance satisfies \(p\le1/(8C_0)\) has polynomial regret on the instance \((\delta_{p/2},\mathrm{Bernoulli}(p))\). The budget has no range term and charges a deterministic competitor its allowance alone, and this fixes the boundary of the class. Since \(-\)Bernoulli\((p)\) is a translate of Bernoulli\((1-p)\), it fails \(\mathrm{VCV}(a,q)\) when \(1-p\le q\), although it is truncated non-positive in the sense of \citet{genalti2024}. Conversely \(N(3,1)\) belongs to \(\mathrm{VCV}(\sqrt{3/2},1/4)\) but is not truncated non-positive. The two assumptions are therefore incomparable, while raw visibility contains every truncated non-positive law, as shown below.

\textbf{Raw visibility ratio.} Write \(\rho(M)=b(M)^+/s(M)\), with value zero when its numerator vanishes and infinity when only its denominator vanishes. This convention covers zero retained energy. For \(X\sim\mathrm{Pareto}(\alpha,x_0)\), with survival function \((x_0/x)^\alpha\) on \(x\ge x_0\) and \(1<\alpha<2\), and \(y=M/x_0\ge1\),
\[
\rho(M)=
\frac{\alpha}{\alpha-1}
\frac{1-y^{-1}}{1+\sqrt{\alpha(1-y^{\alpha-2})/(2-\alpha)}}.
\]
At \(\alpha=3/2\) its maximum is \(\approx1.13018\), above its limit \(\approx1.09808\). The variance-core ratio \(b_E/\sqrt{pv_E}\) along the sets \(E=\{X\le t\}\) gives a lower bound on the required \(a\) (\Cref{fig:visibility}a).

For a finite raw second moment \(v_0=\mathbb E[X^2]>0\), absolute continuity gives \(q_\nu>0\) such that every event \(F\) of probability at most \(q_\nu\) satisfies
\[
\mathbb E[X^2\mathbf1_F]\le v_0/2,\qquad
\mathbb E[(X-\mu)^2\mathbf1_F]\le v_0/2.
\]
Taking \(F=\{|X|>M\}\) and applying Cauchy-Schwarz gives \(b(M)^+\le\sqrt{p(M)V(M)}\). If \(v_0=0\), the law is a point mass at zero. Thus every finite-variance law is eventually raw visible. Apply this fact to \(X-\mu\) and \Cref{app:converse} to obtain eventual VCV, with a law-dependent threshold.

A law is truncated non-positive if \(I_>=\mathbb E[X\mathbf1_{\{|X|>M\}}]\le0\) for every \(M\ge0\). With \(I_\le=\mathbb E[X\mathbf1_{\{|X|\le M\}}]\) and \(p=p(M)\),
\[
b(M)=(1-p)I_>-pI_\le\le p|I_\le|\le p\sqrt{(1-p)V(M)}\le\tfrac12\sqrt{pV(M)}.
\]
This proves raw VT\((1/2,1/2)\). At threshold zero, dominated convergence supplies the same inequality.

Suppose the positive tail is regularly varying with index \(\alpha\in(1,2)\) and the negative tail is negligible relative to it. Karamata's theorem \citep{bingham1987} and tail integration give
\[
p(M)\sim M^{-\alpha}\mathsf L(M),\quad
b(M)\sim\frac\alpha{\alpha-1}M^{1-\alpha}\mathsf L(M),\quad
V(M)\sim\frac\alpha{2-\alpha}M^{2-\alpha}\mathsf L(M).
\]
The negative contributions are negligible by the same integral estimates, and \(\mu p(M)\) is of smaller order. Therefore
\[
\rho(M)\longrightarrow
\frac{\alpha/(\alpha-1)}{1+\sqrt{\alpha/(2-\alpha)}}.
\]
Any strictly larger constant holds beyond some threshold. Choosing a smaller positive omitted-mass bound excludes all smaller thresholds. This threshold depends on the law through its slowly varying factor, so the resulting visibility constants are law dependent.

For the worked Pareto law with scale one, direct integration gives
\[
\begin{aligned}
p(M)&=M^{-\alpha},&
b(M)&=\frac\alpha{\alpha-1}(M^{1-\alpha}-M^{-\alpha}),\\
V(M)&=\frac\alpha{2-\alpha}(M^{2-\alpha}-1),&&M\ge1.
\end{aligned}
\]
For the hidden-mass law \((1-p)\mathrm U\{-1,1\}+p\delta_x\), \(x>1\) and \(p\le1/2\), thresholds \(1\le M<x\) give
\[
\mu=px,\quad b(M)=px(1-p),\quad V(M)=1-p,\quad
\rho(M)=\frac{x\sqrt{p(1-p)}}{1+M\sqrt{p/(1-p)}}.
\]
This expression decreases in \(M\). Thresholds below one omit all observations, and those at least \(x\) omit none. Hence
\[
\frac{x\sqrt p}{2\sqrt2}\le\sup_{M:p(M)\le p}\rho(M)=\rho(1)\le x\sqrt p.
\]
These identities establish the examples exactly.

For a symmetric mean-zero law, let \(F=\{x>0:x\in E^c,-x\in E\}\). Symmetry gives \(b_E\le\int_Fx\,d\nu\), and reflection puts \(-F\) inside \(E\) with mass at most \(p\). Consequently
\[
b_E^2\le p\mathbb E[X^2\mathbf1_E]
=p\left(v_E+\frac{b_E^2}{1-p}\right),\qquad
b_E^2\le p\frac{1-p}{1-2p}v_E\le\tfrac32pv_E
\]
when \(b_E>0\) and \(p\le1/4\). This proves the uniform symmetry claim. A raw constant also follows directly. Write \(X=\mu+\zeta\) with symmetric \(\zeta\). If \(p(M)\le1/4\), then \(M\ge|\mu|\). For \(\mu>0\), cancellation gives
\[
b(M)=\mathbb E[\zeta\mathbf1_{\{M-\mu<\zeta\le M+\mu\}}]\le(M+\mu)p(M),\qquad
V(M)\ge\mu^2(1/2-p(M)).
\]
Thus \(\mu p(M)\le\sqrt{V(M)p(M)}\) and \(b(M)\le s(M)\). For \(\mu<0\), the cancellation has the opposite sign and \(b(M)\le0\). For \(\mu=0\), it vanishes. Hence symmetry about an arbitrary mean implies raw VT\((1,1/4)\) as well.

For \(X\sim\mathrm{Pareto}(\alpha,x_0)\), the mean is \(\mu_P=\alpha x_0/(\alpha-1)\). If the mean-centered omitted probability is at most \(q_\alpha=[(\alpha-1)/(2\alpha)]^\alpha\), its threshold satisfies \(M\ge\mu_P\). Only the right tail remains, and
\[
b(M)=p(M)\left[\frac\alpha{\alpha-1}M+\frac{\mu_P}{\alpha-1}\right]
\le\frac{\alpha+1}{\alpha-1}Mp(M).
\]
For \(\alpha\in[\underline\alpha,2)\) with \(\underline\alpha>1\), the constants \(c=(\underline\alpha+1)/(\underline\alpha-1)\) and \(q=[(\underline\alpha-1)/(2\underline\alpha)]^2\) hold uniformly. \Cref{prop:threshold} then yields uniform VCV constants over translations and scales. For \(\underline\alpha=3/2\) we have \(c=5\) and \(q=1/36\), which give VCV\((34\sqrt2,1/2313)\). For the two-atom family \(\Pi_p\) of \Cref{app:raw}, at raw thresholds between its two atoms,
\[
b=1-p,\qquad V=(1-p)/p,\qquad s(M)\ge\sqrt{1-p}\ge b.
\]

\section{Translation experiment}\label{app:translation}

\begin{figure}[!ht]
\centering
\includegraphics[width=\textwidth]{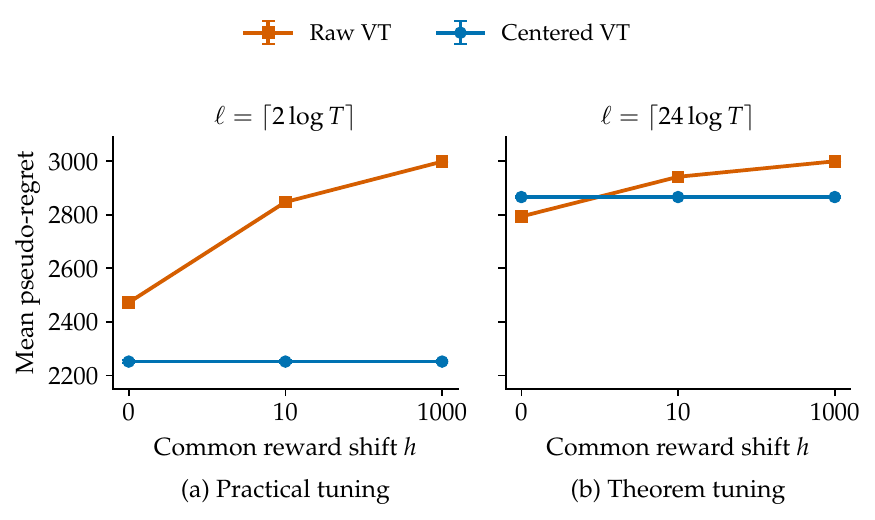}
\caption{Translation sensitivity. Mean pseudo-regret with 95 percent Monte Carlo intervals based on Student's \(t_{199}\) quantile. The practical setting is uncertified. Some intervals are smaller than the plotting symbols. Common shifts preserve centered VT's action sequence.}
\label{fig:translation}
\end{figure}

\Cref{fig:translation} and \Cref{tab:translation} compare raw and centered VT-UCB (raw VT and centered VT below) on \(N(3+h,1)\) and \(N(2.7+h,1)\) at \(T=20{,}000\), with 200 independent replications, base seed 42, and common random numbers across methods and shifts. Both use \((a,q)=(\sqrt{3/2},1/4)\). Centering reserves 319 pilot observations per arm. The theorem setting uses \(\ell=238\) and 1,904 fresh warm-up observations per arm. A practical setting uses \(\ell=20\) and 160 fresh observations, outside the range of the theorem.

\begin{table}[!ht]
\centering
\begin{tabular}{llrrr}
\toprule
Setting & Method & \(h=0\) & \(h=10\) & \(h=1000\)\\
\midrule
Theorem, \(\ell=238\) & raw VT & 2793.3 & 2941.3 & 2999.1\\
 & centered VT & 2865.7 & 2865.7 & 2865.7\\
Practical, \(\ell=20\) & raw VT & 2472.3 & 2847.3 & 2997.9\\
 & centered VT & 2252.1 & 2252.1 & 2252.1\\
\bottomrule
\end{tabular}
\caption{Mean pseudo-regret over 200 replications at \(T=20{,}000\). The largest 95 percent half-width is \(6.5\).}
\label{tab:translation}
\end{table}

\FloatBarrier
Two reference points calibrate these values. Uniform play has regret \(T\Delta/2=3000\), and UCB with known unit variance and bonus \(\sqrt{2\log t/n}\) has mean regret \(53.0\) over 600 replications on the unshifted instance, with 95 percent interval half-width \(2.0\). Every configuration in \Cref{tab:translation} pulls the worse arm at least 37 percent of the time. Even at sample size \(T\), the certified width \(3(1+a)\widehat w\) of a unit-variance arm is \(\approx2.78\) with \(\ell=238\) and \(\approx0.49\) with \(\ell=20\), against a gap of \(0.3\).

All runs remain in the pre-logarithmic regime, where the certified widths exceed the gap, so the experiment isolates the effect of translation. Centered VT has regret constant in the shift, because its action sequence is translation invariant, and raw VT has regret increasing in the shift. At zero shift under theorem tuning, raw VT has the lower regret, \(2793.3\) against \(2865.7\).

\end{document}